\documentclass[10pt,twocolumn,letterpaper]{article}

\usepackage[pagenumbers]{cvpr} 

\usepackage{graphicx}
\usepackage{amsmath}
\usepackage{amssymb}
\usepackage{booktabs}
\usepackage{mathtools}
\usepackage{bbm}
\usepackage{algorithm}
\usepackage{algorithmic}
\usepackage{multirow}
\usepackage[accsupp]{axessibility}

\usepackage[pagebackref,breaklinks,colorlinks]{hyperref}

\usepackage[capitalize]{cleveref}
\crefname{section}{Sec.}{Secs.}
\Crefname{section}{Section}{Sections}
\Crefname{table}{Table}{Tables}
\crefname{table}{Tab.}{Tabs.}

\def\cvprPaperID{6763} 
\def\confName{CVPR}
\def\confYear{2022}

\newcommand{\bs}[1]{\boldsymbol{#1}}
\newcommand{\mbf}[1]{\mathbf{#1}}
\newcommand{\mbb}[1]{\mathbb{#1}}
\newcommand{\mca}[1]{\mathcal{#1}}
\newcommand{\indep}{\perp \!\!\! \perp}
\newcommand{\rep}[2]{\bs{r}_{#1}^{#2}}  
\newcommand{\repf}[2]{\mbf{r}_{#1}^{#2}} 

\begin{document}


\title{Contrastive Conditional Neural Processes}

\author{
Zesheng Ye \qquad \qquad Lina Yao \\
{\small The University of New South Wales} \\
{\tt \small zesheng.ye@unsw.edu.au \qquad lina.yao@unsw.edu.au}
}

\maketitle

\begin{abstract}
    Conditional Neural Processes~(CNPs) bridge neural networks with probabilistic inference to approximate functions of Stochastic Processes under meta-learning settings.
    Given a batch of non-{\it i.i.d} function instantiations, CNPs are jointly optimized for in-instantiation observation prediction and cross-instantiation meta-representation adaptation within a generative reconstruction pipeline.
    There can be a challenge in tying together such two targets when the distribution of function observations scales to high-dimensional and noisy spaces.
    Instead, noise contrastive estimation might be able to provide more robust representations by learning distributional matching objectives to combat such inherent limitation of generative models.
    In light of this, we propose to equip CNPs by 1) aligning prediction with encoded ground-truth observation, and 2) decoupling meta-representation adaptation from generative reconstruction.
    Specifically, two auxiliary contrastive branches are set up hierarchically, namely in-instantiation temporal contrastive learning~({\tt TCL}) and cross-instantiation function contrastive learning~({\tt FCL}), to facilitate local predictive alignment and global function consistency, respectively.
    We empirically show that {\tt TCL} captures high-level abstraction of observations, whereas {\tt FCL} helps identify underlying functions, which in turn provides more efficient representations.
    Our model outperforms other CNPs variants when evaluating function distribution reconstruction and parameter identification across 1D, 2D and high-dimensional time-series.
\end{abstract}

\section{Introduction}
\label{sec:intro}

Supervised generative models learn to reconstruct joint distribution with certain prior incorporated to achieve parameter estimation.
This brings in a natural fit to meta-learning~\cite{vilalta2002perspective,chen2017learning, hospedales2020meta} paradigm, where knowledge acquired from previous tasks can help fast adaptation in solving novel tasks drawn from the same task distribution. Conditional Neural Processes~\cite{garnelo2018conditional, kimanp18, Gordon2019ConvCNP} (CNPs) lie at the intersection between these two.
Mathematically, for $T$ steps of a function instantiation $D = \{ (x_{t}, \bs{y}_{t}) \}_{t=1}^{T}$ sampled from a time-series, where each time index $x$ has an associated observation $y \subseteq \mbb{R}^{d}$, CNPs learn to reconstruct the observation of a given query index $x_{q}$, with a set of time-observation pairs $\mathcal{C} \subseteq D$ as input.
Specifically, CNPs are built upon an encoder-decoder, where the encoder $h : \mathcal{C} \rightarrow \mbf{r}_{C}$ summarizes all the input as a contextual representation $\mbf{r}_{C}$.
The decoder takes $\mbf{r}_{C}$ and query index $x_{q}$ and outputs estimated observation $g: \mbf{r}_{C}, x_{q} \rightarrow \hat{\bs{y}}_{q} \approx \bs{y}_{q}$.
In a meta-learning context, CNPs may deal with some non-{\it i.i.d} function instantiations.
If, for instance, given two function instantiations $D_{1} \sim f_{1} = 5x + 10$ and  $D_{2} \sim f_{2} = 8x + 1$, the representation derived from $f_{1}$ cannot be used directly to reconstruct observations of $f_{2}$, but rather requires a cross-instantiation adaption step for meta-representation.
In CNPs, the adaptation takes place implicitly along with data reconstruction.

The Bayesian principles enable CNPs to quantify uncertainties while performing predictive tasks under limited data volume.
Moreover, CNPs are designed to model a distribution over functions whereas conventional deep learning models do not, suggesting better generalization when multiple functions are in play.
These properties are desirable in applications like traffic forecasting, trajectory prediction, and activity recognition~\cite{ivanovic2019trajectron, salzmann2020trajectron++, messing2009activity}, where observations might be corrupted or sampled from different data-generation functions.

The CNP's variants attempt to improve the encoder $h$ by introducing a range of inductive biases~\cite{kimanp18, Gordon2019ConvCNP, KawanoKSIM21} to learn appropriate structures from data.
Despite the performance gains achieved, the inherent generative nature of these models has raised some concerns, especially when they are considered together: 1) Correlations between observations are not modeled in CNPs. While introducing a global latent variable can yield more coherent predictions~\cite{garnelo2018neural, kimanp18, Foong2020ConvNP}, these attempts are also fraught with intractable likelihood; 2) High-dimensional observations challenge the capacity of CNPs since generative models reconstruct low-level details but hardly form high-level abstractions~\cite{DBLP:conf/iclr/KipfPW20, Racah2020Slot}; and 3) Supervision collapse in meta-learning~\cite{doersch2020crosstransformers} occurs when prediction and transfer tasks are entangled in CNPs~\cite{gondal2021function}.

\begin{figure}
    \centering
    \includegraphics[width=0.8\linewidth]{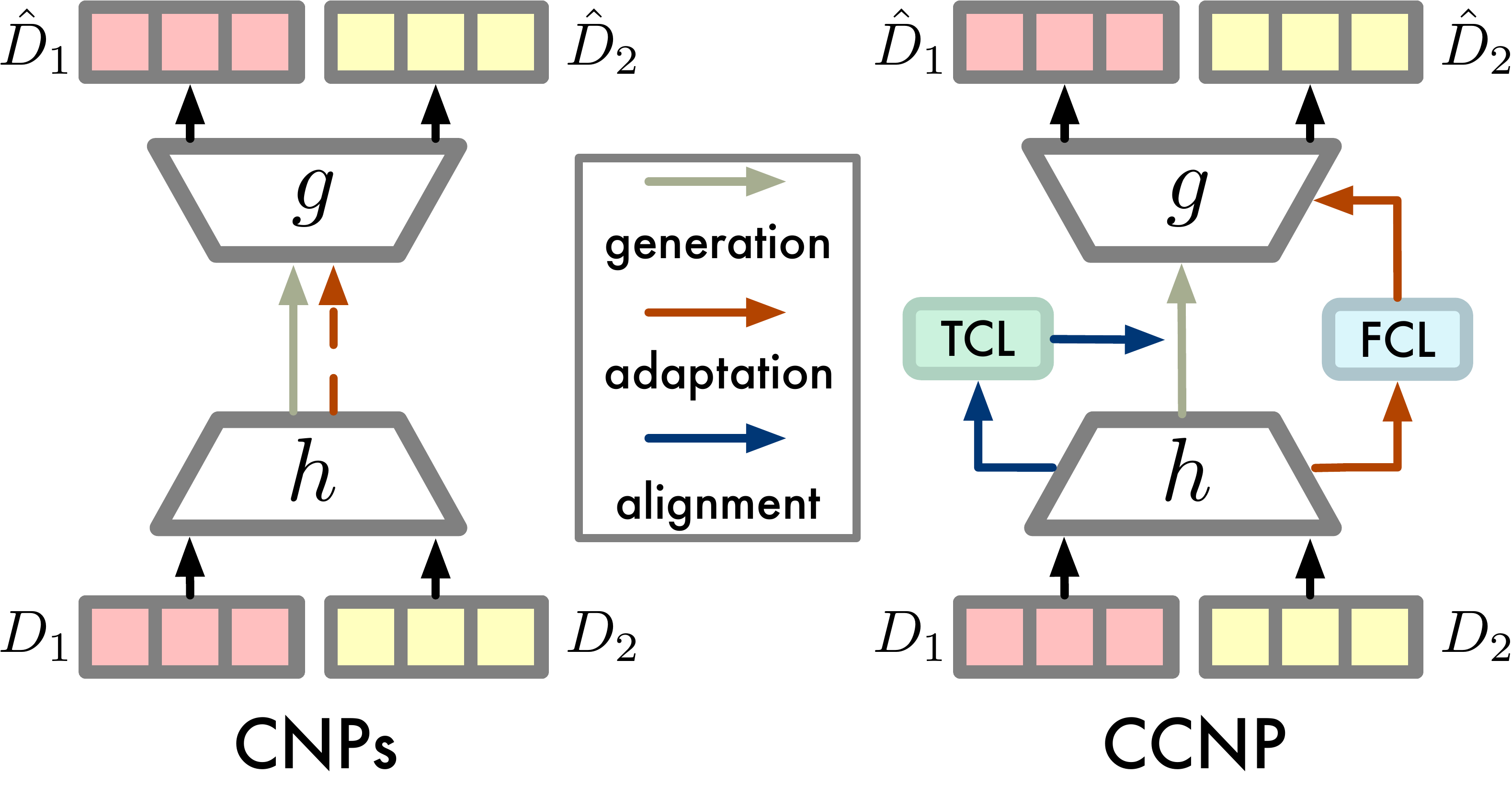}
    \caption{The motivation of the proposed CCNP against CNPs. While adaptation is {\bf implicitly} done within generation in CNPs, CCNP {\bf explicitly} processes it with {\tt FCL} branch, and {\tt TCL} branch ensures the alignment of high-level abstraction of observations.}
\end{figure}

To address this, our first motivation centers on decoupling model adaptation from generative reconstruction through explicit contrastive learning, with the aim of making each objective attends to its own duties.
Contrastive learning has become increasingly popular thanks to its clearer (discriminative) supervisions.
Despite being implemented differently in a range of applications~\cite{chen2020simple, gao2021simcse, you2020graph}, the shared principle behind these contrastive methods is to create different types of tuples by utilizing certain structures in the data and training the model to identify the types.
For instance in SimCLR~\cite{chen2020simple}, images are firstly transformed randomly with two augmentations, which are then formed as contrastive pairs labeled as either positive or negative, depending on whether two augmentations are from the same instance.
Likewise, contrastive learning has renewed a surge of interest in conditional density estimation\cite{pmlr-v130-ton21a, pacchiardi2022score, tsai2022conditional}.
In this scope, discriminative pretext tasks are designed to create more efficient representations for modeling higher-dimensional data than generative counterparts~\cite{liu2021self}.
One should however note that they do so at the expense of flexibility, as only downstream tasks that are directly related to the contrastive task can be accommodated.

As such, we investigate a generative-contrastive model in an effort to combine the complementary benefits of both approaches while mitigating the deficiencies of each.
Existing studies linking contrastive learning to generative models spotlight the "pre-train then fine-tune" scheme~\cite{winkens2020contrastive, tack2020csi}.
Contrary to this, we aim to train a hybrid model in an end-to-end way, without losing the flexibility of generative models, as well as the efficiency of contrastive models.

To this end, we propose Contrastive Conditional Neural Process~(CCNP) that extends generative CNPs with two auxiliary contrastive branches, coined {\it in-instantiation temporal contrastive learning}~({\tt TCL}) and {\it cross-instantiation function contrastive learning}~({\tt FCL}).
A particular emphasis is placed on the use of contrastive objectives to handle noisy high-dimensional observations.
Concretely, {\tt TCL} imposes {\it local} alignment between predictive representations and encoded ground-truth observations of the same timestep to model temporal correlations, enhancing the model's scalability to higher dimensions.
FCL encourages {\it global} consistency across different partial views of the same instantiation to separate adaptation from reconstruction, thus improving the transferability of meta-representations.


\

\noindent {\bf Contributions.}
1)~We present an end-to-end generative-contrastive meta-learning model based on CNPs.
2)~We demonstrate that incorporating proper contrastive objectives into generative CNPs contributes to complementary benefits, and hopefully leads to new avenues for research in CNPs.
3)~We empirically verify CCNP outperforms other CNP baselines in terms of function distribution reconstruction and parameter identification across diverse tasks, including 1D few-shot regression, 2D population dynamics prediction and high-dimensional sequences reconstructions.


\section{Related Works}
\label{sec:rel}

\noindent {\bf Conditional Neural Processes.}
Conditional Neural Process~(CNP)~\cite{garnelo2018conditional} was proposed to perform function approximation under a meta-learning setting.
Meta-learning refers to a learning paradigm that facilitates rapid model adaptation across multiple related tasks.
This is typically implemented via a bi-level learning setting, where the model solves predictive tasks within inner-level loop, while optimizing the generalization ability during outer-level learning.
CNPs wrap up the inner step with outer-level optimization based on a simple encoder-decoder architecture.
Recent progress includes baking a variety of inductive biases into the model so that the corresponding structure can be learned from data.
AttnCNP~\cite{kimanp18} uses attention~\cite{VaswaniAttention17} to improve interpolation performance inside the range of training input.
ConvCNP~\cite{Gordon2019ConvCNP} realizes translation-equivariance and enhances extrapolation capability outside the training range.
The latest evolution involves imposing symmetries on CNPs, \eg group equivariance in~\cite{KawanoKSIM21} and SteerableCNP~\cite{holderrieth2021equivariant} to ensure spatial invariance.
Even so, the majority of their efforts are devoted to encoder while neglecting supervision collapse associated with the inherent limitation of generative reconstruction.
ContrNP~\cite{kallidromitis2021contrastive} shares the motivation of preserving function consistency with us, the proposed CCNP is more scalable owing to the {\tt TCL} branch.

\begin{figure*}[h]
    \centering
    \includegraphics[width=\textwidth]{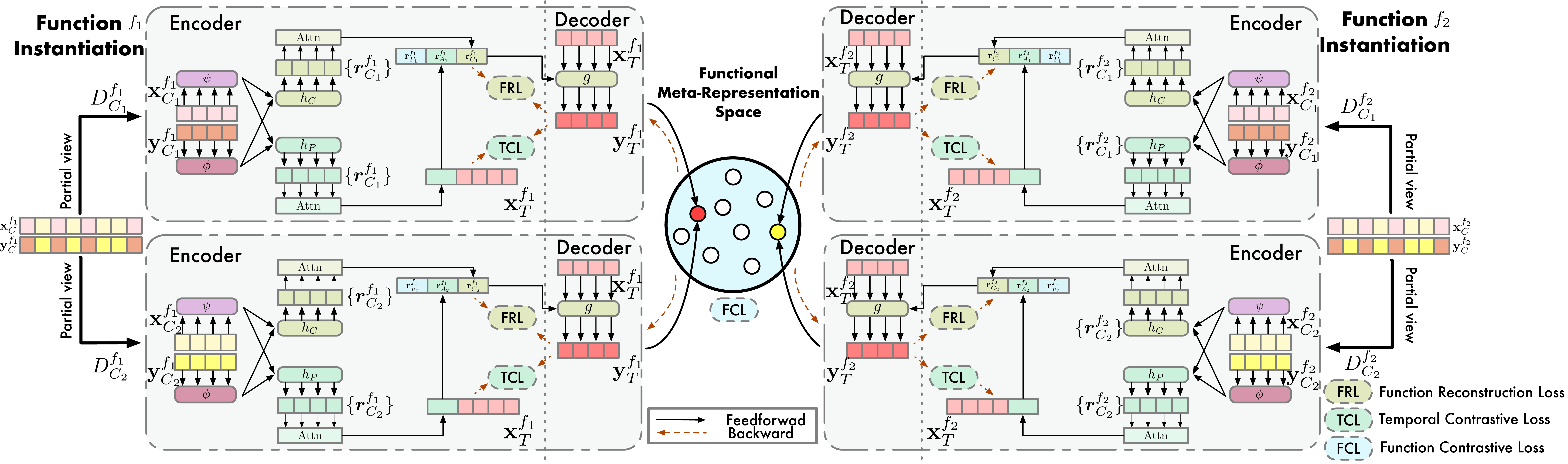}
    \caption{The proposed Contrastive Conditional Neural Processes. We showcase the scenario where there are two function instantiations (\ie $f_{1}$ and $f_{2}$) with two partial views (\ie $D_{C_{1}}^{f}$ and $D_{C_{2}}^{f}$) of each $f$ for illustration only.
    Each instantiation $f$ is involved with three losses, including a) Function Reconstruction Loss~{\tt FRL}, implemented with an encoder-decoder generative module; b) in-instantiation Temporal Contrastive Loss~{\tt TCL} enforcing the predictive function embedding to align with ground-truth function embedding; and c) cross-instantiation Function Contrastive Loss~{\tt FCL} that regularizes the transferability of meta-representation for each instantiation.
    }
    \label{fig:overall}
\end{figure*}

\

\noindent {\bf Contrastive Conditional Density Estimation.}
The common strategy held by~\cite{pmlr-v130-ton21a, gondal2021function, pacchiardi2022score, tsai2022conditional} is to bypass likelihood-based distribution inference with noise contrastive estimation~(NCE)~\cite{gutmann2010noise}.
MetaCDE~\cite{pmlr-v130-ton21a} learns a mean embeddings to estimate conditional density of multi-modal distribution.
FCLR~\cite{gondal2021function} replaces the decoder of CNPs with self-supervised contrastive signals to model function representations.
CReSP~\cite{mathieu2021contrastive} takes a further step upon FCLR and forms a semi-supervised framework to learn representations at specific indices of time-series.
While both FCLR and CReSP also seek for preserving function consistency as in ContrNP and CCNP, they stand for different tasks.
CReSP primarily accommodates downstream tasks without relying on reconstruction, whereas ContrNP and CCNP are intended to reconstruct observations ultimately.

\noindent {\bf Summary.}
Our work is orthogonal to recent signs of progress in either branch, yet combines the merits of both into one framework.
To CNPs, we strip off model adaptation from the function generation process, and set up a contrastive likelihood-free objective in its place.
Moreover, we learn to extract the high-level abstraction of function observations rather than wasting model's capacity on reconstructing every low-level detail.
Such the hierarchical contrastive objectives aid in scaling the model to handle high-dimensional observations and to provide robust meta-representation across multiple non-{\it i.i.d} instantiations.
To CDEs, we investigate a hybrid generative-contrastive meta-learning model while the previous ones doing away from the reconstruction.
Therefore, the proposed method may serve as a universal plug, and being able to be combined with any instance from the CNP family.

\section{Contrastive Conditional Neural Processes}
\label{sec:method}
\cref{fig:overall} overviews CCNP, wherein the model is tasked with optimizing three objectives, including a) generative function observation reconstruction loss~({\tt FRL}) (\cref{sec:methods_rfd}), plus b) in-instantiation temporal contrastive alignment loss~({\tt TCL}) (\cref{sec:methods_tcl}) and c) cross-instantiation function contrastive identification loss~({\tt FCL}) (\cref{sec:methods_fcl}).


\subsection{Preliminaries}
\label{sec:method_problem}
\noindent {\bf Problem Statement.}
We detail meta-learning for time-series prediction as modeling stochastic processes (SPs) $\mca{F} : \mca{X} \rightarrow \mca{Y}
$, with $\mca{X} = [0, +\infty)$ is $1$-dimensional scalar time indices and $\mca{Y} \subseteq \mathbb{R}^{d}$ represents $d$-dimensional observations.
Let $D_{C}^{f} = \{ (x_{c}^{f}, \bs{y}_{c}^{f}) \}_{c \in I_{C}^{f}} $ be a set of observed sample pairs uniformly drawn from a specific underlying data-generating function $f \sim \mca{F}$, indexed by $I_{C}^{f}$, referred to the context set of functional instantiation $f$.
Each $D_{C}^{f}$ is associated with a learning task that maps context set to a contextual representation $\repf{C}{f}$ defined over $f$, and reconstruct the function observations on a superset of context set, denoted by the target set $D_{C}^{f} \subset D_{T}^{f}= \{ (x_{t}^{f}, \bs{y}_{t}^{f}) \}_{t \in I_{T}^{f}}$.

To model the variability of SPs with meta-learning, predictions at target indices $\mbf{x}_{T}^{f}$ are specified as predictive posterior distribution $p(\mbf{y}_{T}^{f} | \mbf{x}_{T}^{f}; D_{c}^{f})$ that are consistent with, say, a set of $F$ function instantiations in practice (\ie $f \in F \sim \mca{F}$), conditioning on corresponding context set.
Assume, any finite sets of function evaluations of $f$ are conditionally independent, and jointly Gaussian distributed given context set, a generative model is obtained where the output is a conditional predictive distribution
\begin{align}\label{eq: cnp_llh_1}
    p(\mbf{y}_{T}^{f} | \mbf{x}_{T}^{f} ; D_{C}^{f} ) & = \prod_{t=1}^{|D_{T}^{C}|} p(\bs{y}_{t}^{f} | \bs{\Phi}(D_{C}^{f})(\bs{x}_{t}^{f})) \nonumber \\
                                                      & = \prod_{t=1}^{|D_{T}^{C}|} \mca{N} (\bs{y}_{t}^{f} ; \bs{\mu}_{t}^{f}, \bs{\Sigma}_{t}^{f})
\end{align}
\begin{flalign*}
     & \mathrm{with}~(\bs{\mu}_{t}, \bs{\Sigma}_{t}) = \bs{\Phi}(D_{C}^{f})(\bs{x}_{t}) = g(h(D_{C}^{f}), \bs{x}_{t}^{f}) = g(\repf{C}{f}, \bs{x}_{t}^{f}) \\
     & \mathrm{and}~f(\bs{x}_{t}^{f}) \indep f(\bs{x}_{t^{\prime}}^{f}) \mid D_{C}^{f}, \;\; t \neq t^{\prime} \in I_{T}^{f}
\end{flalign*}
where $\bs{\Phi} = \{ h, g \}$ is implemented as encoder-decoder that approximates the function mapping from a dataset of context set to predictive distributions over arbitrary target sets.

\

\noindent {\bf Workflow in CNPs}.
Model generation in CNPs starts with passing through every context sample in $D_{C}^{f}$ into encoder network $h$ to acquire a finite-dimensional representation, and performing permutation-invariant aggregation over whole context set $\repf{C}{f} = \square_{c \in \mca{C}}(h(x_{c}^{f}, \bs{y}_{c}^{f}))$.
We note that $\square(h(\cdot, \cdot))$ is implemented with a {\it DeepSet}~\cite{ZaheerKRPSS17} structure, where the input are arranged as a set to ensure operation $\square(\cdot)$ is invariant to various input orders and size.
The decoder network $g$ then takes the conditional prior $\rep{C}{f}$ for target indices to produce a posterior distribution
$p(\mbf{y}_{T}^{f} | \mbf{x}_{T}^{f} ; D_{C}^{f})$, of which the predictive performance is measured via log-likelihood estimation on the target set.

\subsection{Reconstructing Function Distribution}
\label{sec:methods_rfd}
\noindent {\bf Encoder}.
The function observation reconstrution objective {\tt FRL} is built upon CNPs.
Unlike vanilla CNP that concatenates each pair of $\bs{x}_c^{f}$ with $\bs{y}_c^{f}$ to the encoder $h_{C}$, we separately pre-process $\bs{x}_c^{f}$ and $\bs{y}_c^{f}$ using diverse feature extraction networks $\psi(\bs{x}^{f})$ and $\phi(\bs{y}^{f})$.
Considering that $\bs{y}^{f}$ could be in a high-dimensional setting, \ie $\mbb{R}^{d \geq 3}$, domain-specific feature extractors can be applied to obtain efficient embeddings of observation, \eg CNN for images and GNN for graph-structured data.
Following that, the encoder $h_{C}(\cdot, \cdot)$ is applied to each time-observation pair of the context set $D_{C}$ to obtain the corresponding {\it local} representation $\rep{c}{f}$
\begin{equation}\label{eq: representation_2}
    \rep{c}{f} = h_{C} \left(\psi \left(x_{c}^{f} \right), \phi \left(\bs{y}_{c}^{f} \right) \right)
\end{equation}

\noindent {\bf Position-Aware Aggregation}.
CNP~\cite{garnelo2018conditional} applies mean aggregation over the whole context set, which, however ignores the relative position between any two context indices and results in under-fitting~\cite{kimanp18, Gordon2019ConvCNP}.
To capture such correlation, we pass encoded representations $\rep{c=1:|D_{C}|}{f}$ through a multi-head self-attention~\cite{VaswaniAttention17} module to calculate the position-aware importance matrix, followed by mean aggregation, to compute the contextual representation that summarizes all the time-observation pairs from $D_{C}$
\begin{equation}\label{eq: attn_3}
    \repf{C}{f} = \frac{1}{|D_{C}^{f}|} \sum_{ ( x_{c}^{f}, \bs{y}_{c}^{f}) \in D_{C}^{f} } \sum_{k} \frac{e^{\mbf{W}_{q} \rep{c}{f} \cdot \mbf{W}_{k} \rep{k}{f} / \sqrt{d}}}{\sum_{k} e^{\mbf{W}_{q} \rep{c}{f} \cdot \mbf{W}_{k} \rep{k}{f} / \sqrt{d}}} \, \mbf{W}_{v} \rep{k}{f}
\end{equation}
where $d$ is the dimension of $\rep{c}{f}$, $\mbf{W}_q, \mbf{W}_k, \mbf{W}_v $ are linear operators applied to $\rep{c}{f}$ and $\rep{k}{f}$ for $k = 1:|D_{C}^{f}|$.

\noindent {\bf Decoder}.
\label{methods:decoder}
We concatenate the query index $\psi(\bs{x}_{t}^{f})$ with contextual representation $\repf{}{f}$, and use decoder $g$ to estimate observation $\hat{\mbf{y}}_{T}^{f}$, where $\repf{}{f}$ is a non-linear transformation of $[\repf{C}{f}, \repf{P}{f}, \repf{F}{f}]$.
The latter two are obtained by following the same procedure as~\cref{eq: representation_2},~\cref{eq: attn_3} with different encoder $h_{P}$ and $h_{F}$, and are optimized according to~\cref{sec:methods_tcl} and~\cref{sec:methods_fcl}, respectively.
Following the common empirical evaluation criteria of CNPs~\cite{le2018empirical}, reconstructed observations of the target set $D_{T}$ is assumed to be a joint Gaussian distribution.
\begin{align}\label{eq: cnp_predict_4}
    \hat{\mbf{y}}_{T}^{f}              & \sim \log \prod_{t=1}^{|D_{T}^{f}|} \mca{N} (\bs{\mu}_{t}, \bs{\sigma}_{t}) \\
    ~\mathrm{with}~       \bs{\mu}_{t} & = \mathrm{softmax}(\mu(\rep{t}{f})), \nonumber                              \\
    \bs{\sigma}_{t}                    & = \mathrm{diag}(0.9\times\mathrm{softplus}(\rep{t}{f}) + 0.1) \nonumber
\end{align}
where $\mu(\cdot), \sigma(\cdot)$ are linear transformations of target input and output predictive mean and variance values.



\subsection{Aligning Predictive Temporal Representation}
\label{sec:methods_tcl}
\noindent {\bf Likelihood-free Density Estimation}.
We expect the predictive representation to be similar to the encoding of ground-truth observation~\cite{DBLP:conf/iclr/KipfPW20,mathieu2021contrastive}, so as to realize temporal {\it local} alignment
For instance, let $\bs{x}_{t}^{f}$ be the query target index, we attempt to maximize the density ratio of its predictive embedding conditioned on sets of context set $D_{C}^{f \in F}$ sampled from $F$ different instantiations.

\noindent {\bf In-Instantiation Temporal Contrastive Loss}.
In practice, we formulate an in-instantiation temporal contrastive learning~({\tt TCL}) objective, optimized with InfoMax~\cite{oord2019representation}-based loss to estimate such ratio of likelihood.
The concatenated target index and contextual representation is transformed with predictive head $\varphi$ yielding the predictive embedding $\hat{\bs{\varphi}}_{t}^{f}$ at target index $\bs{x}_{t}^{f}$.
We further apply a non-linear projection head $\rho_{p}: \mathbb{R}^{\phi, \varphi} \mapsto \mathbb{R}^{z} $ to map these embeddings to a low-dimensional space for similarity measurement~\cite{chen2020simple}.
\begin{equation}\label{eq: projection_head_6}
    \hat{\bs{z}}_{t}^{f} = \rho_{p}(\hat{\bs{\varphi}}_{t}^{f}) ,  \bs{z}_{t}^{f} = \rho_{p}(\varphi(\bs{y}_{t}^{f})) \; \mathrm{with} \; \hat{\bs{\varphi}}_{t}^{f} = \varphi(\bs{x}_{t}^{f}, \repf{C}{f})
\end{equation}
Since $\hat{\bs{z}}_{t}$ and $\bs{z}_{t}$ is considered as positive pair, function embeddings of the remain indices within the batch become negative samples.
The {\tt TCL} loss can therefore be given by
\begin{align}\label{eq: tcl_7}
     & \mca{L}_{\mathrm{TCL}}= - \sum_{f=1}^{F} \sum_{t=1}^{T} \nonumber             \\
     & \log \frac{ e^{\mathrm{sim}(\hat{\bs{z}}_{t}^{f}, \bs{z}_{t}^{f}) / \tau } }{
    e^{\mathrm{sim}(\hat{\bs{z}}_{t}^{f}, \bs{z}_{t}^{f}) / \tau } +
    \sum_{f^{\prime}}^{F} \sum_{i=1}^{T+C} \mathbbm{1}_{[i \neq t]} e^{ \mathrm{sim}(\hat{\bs{z}}_{t}^{f}, \bs{z}_{i}^{f^{\prime}}) / \tau }
    }
\end{align}
where $\mathrm{sim}(\mbf{a}, \mbf{b}) = \frac{ \mbf{a}^{\top} \mbf{b} }{|| \mbf{a} || \, || \mbf{b} ||}$ is cosine similarity. $\tau$ denotes the temperature coefficient. $C$, $T$ and $F$ refers to the number of sampled context set, target set and observation sequences, respectively.
By minimizing~\cref{eq: tcl_7}, we ensure the predicted embedding is closer to the embedding of the true outcome than embeddings of other random outcomes.

\begin{figure}
    \centering
    \includegraphics[width=\linewidth]{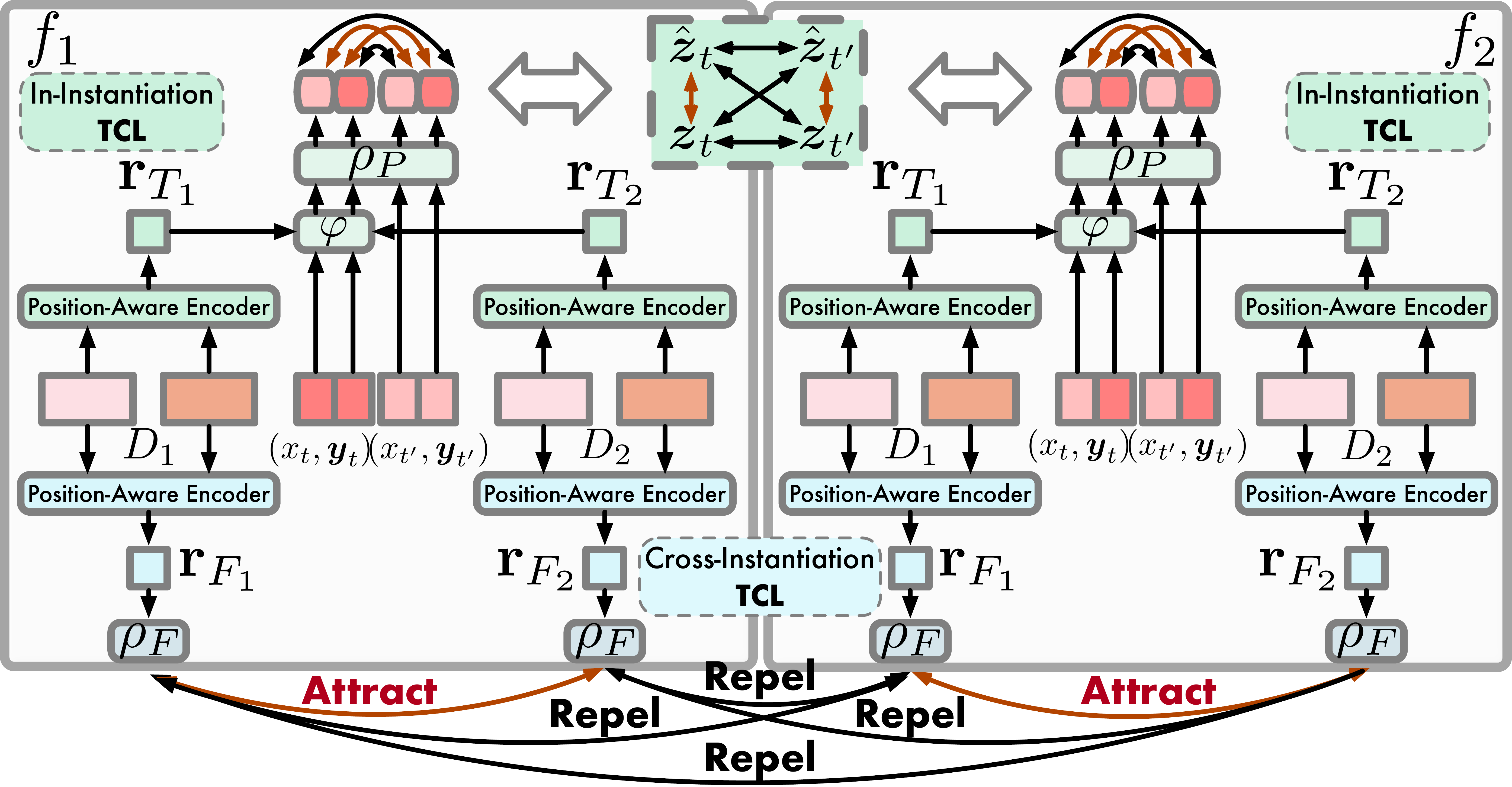}
    \caption{Conceptual illustration of {\tt TCL} and {\tt FCL}.
    We note that details of $\phi, \psi$ are omitted for brevity. The upper half (green) depicts in-instantiation {\tt TCL} with respect to \cref{eq: representation_2},~\cref{eq: attn_3},~\cref{eq: projection_head_6}~\ie the objective is optimized within each instantiation $f$.
    The lower (blue) half discusses the cross-instantiation {\tt FCL} regarding~\cref{eq: representation_2},~\cref{eq: attn_3},~\cref{eq: fcl_split_8},~\cref{eq: fcl_9}.}
    \label{fig:method_contrastive}
\end{figure}

\subsection{Regularizing Functional Meta-Representation}
\label{sec:methods_fcl}

\noindent {\bf Self-Supervised Function Identification.}
We further explicitly generalize contextual representation across instantiations as in~\cite{gondal2021function}, rather than entangle meta-representation acquisition with predictive meta-task together as in CNPs.
Different partial observations from the same instantiation are expected to be identified to the same data-generating function.
By doing so, we encourage the representations of the same partial context set to be mapped closer in the embedding space, while in contrast, different ones should be pushed away.
This motivates the function contrastive learning~({\tt FCL}) objective.
In contrast to {\tt TCL} using target indices and observations, {\tt FCL} solely optimizes the training procedure with only context sets in a self-supervised spirit.

\

\noindent {\bf Cross-Instantiation Function Contrastive Loss}.
With accessing to $F$ context sets drawn from corresponding functions, we halve $D_{C}^{f}$ into two disjoint subsets by randomly selecting $|D_{C}^{f}|/2$ samples to comprise the partial view $D_{C_{1}}^{f}$, then the left half makes up of another partial view $D_{C_{2}}^{f}$, \ie
\begin{equation}\label{eq: fcl_split_8}
    D_{C}^{f} = D_{C_{1}}^{f} \cup D_{C_{2}}^{f} \; \mathrm{with} \; D_{C_{1}}^{f} \cap D_{C_{2}}^{f} = \varnothing
\end{equation}
We process each partial view in the similar way as {\tt TCL} to obtain self-attentive representation $\mbf{r}_{F_{i}}^{f}, \mbf{r}_{F_{j}}^{f}$.
Subsequently, we pass them through the projection head $\rho_{F}: \mathbb{R}^{r} \mapsto \mathbb{R}^{z}$ and get the corresponding low-dimensional representations $\boldsymbol{q}_{i}^{f} = \rho_{F}(\mbf{r}_{F_{i}}^{f})$ and $\boldsymbol{q}_{j}^{f} = \rho_{F}(\mbf{r}_{F_{j}}^{f})$. {\tt FCL} is formulated as
\begin{align}\label{eq: fcl_9}
     & \mca{L}_{\mathrm{FCL}} = - \sum_{f=1}^{F} \sum_{1 \leq i < j \leq 2} \nonumber \\
     & \log \frac{
    e^{\mathrm{sim}(\boldsymbol{q}_{i}^{f}, \boldsymbol{q}_{j}^{f}) / \tau }
    }{
    e^{\mathrm{sim}(\boldsymbol{q}_{i}^{f}, \boldsymbol{q}_{j}^{f}) / \tau } + \sum_{f^{\prime}}^{F} \sum_{1 \leq i \leq j \leq 2} \mathbbm{1}^{[f \neq f^{\prime}]} e^{\mathrm{sim}(\boldsymbol{q}_{i}^{f}, \boldsymbol{q}_{j}^{f^{\prime}}) / \tau}
    }
\end{align}
\begin{algorithm}
    \caption{Episodic Training for CCNP}
    \label{algo:training}
    \begin{algorithmic}[1]
        \REQUIRE A dataset of $F$ function instantiations $\{ D^{f} \}_{f \in F}$.
        \REQUIRE context\_range $ = C $, extra\_target\_range $ = T$.
        \ENSURE Parameters $ \Theta =\{ \phi, \psi, h_{C}, h_{P}, h_{F}, g, \rho_{F}, \rho_{P}, \varphi \}$
        \\ /* {\it Sampling context set and target set} */
        \FOR{$f \in F$}
        \STATE Sample context size $|D_{C}^{f}| \leftarrow C$~with~$C \sim \mca{U}(1, C)$
        \STATE Sample target size $|D_{T}^{f}| \leftarrow C + T$with$T\sim \mca{U}(1, T)$
        \STATE Uniformly sample context set $D_{C}^{f}$ with size $|D_{C}^{f}|$
        \STATE Uniformly sample target set $D_{T}^{f}$ with size $|D_{C + T}^{f}|$
        \ENDFOR
        \\ /* {\it Optimizing {\tt FCL} Objective} (Section~\cref{sec:methods_fcl}) */
        \STATE Obtain representations of partial views $\repf{F_{i}}{f}, \repf{F_{j}}{f} \leftarrow $~\cref{eq: representation_2},\cref{eq: attn_3},\cref{eq: fcl_split_8},~{\bf foreach}~$f \in F$.
        \STATE Compute $\mca{L}_{\mathrm{FCL}} \leftarrow$~\cref{eq: fcl_9}
        \STATE $\theta \leftarrow \theta - \beta \nabla_{\theta} \mca{L}_{\mathrm{FCL}}$~{\bf foreach}~$\theta \in \{ \psi, \phi, h_{F}, g, \rho_{f} \}$
        \\ /* {\it Optimizing {\tt TCL} Objective}~(\cref{sec:methods_tcl}) */
        \FOR{$f \in F$}
        \STATE Obtain representation $\repf{P}{f} \leftarrow $~\cref{eq: representation_2},~\cref{eq: attn_3}
        \STATE Obtain predictive and ground-truth embeddings of target $\hat{\bs{z}}_{t}^{f}, \bs{z}_{t}^{f} \leftarrow$~\cref{eq: projection_head_6}~{\bf foreach $t \in T$}
        \ENDFOR
        \STATE Compute $\mca{L}_{\mathrm{TCL}} \leftarrow$~\cref{eq: tcl_7}
        \STATE $\theta \leftarrow \theta - \alpha \nabla_{\theta} \mca{L}_{\mathrm{TCL}}$~{\bf foreach}~$\theta \in \{ \psi, \phi, h_{T}, \rho_{P}, \varphi \}$
        \\ /* {\it Optimizing FRL Objective}~(\cref{sec:methods_rfd}) */
        \STATE Obtain context representation $[\repf{C}{f}, \repf{P}{f}, \repf{F}{f}] \leftarrow $~\cref{eq: representation_2} - \cref{eq: attn_3}
        \STATE Compute $\mca{L}_{\mathrm{FRL}} \leftarrow $~\cref{eq: cnp_predict_4}~{\bf foreach}~$f \in F$
        \STATE $\theta \leftarrow \theta - \nabla_{\theta} \mca{L}_{\mathrm{FRL}}$~{\bf foreach}~$\theta \in \{ \psi, \phi, h_{C}, g \}$
    \end{algorithmic}
\end{algorithm}
where $\mathrm{sim}(\mbf{a}, \mbf{b}) = \frac{ \mbf{a}^{\top} \mbf{b} }{|| \mbf{a} || \, || \mbf{b} ||}$ refers to the cosine similarity and $\tau$ is the temperature coefficient.

\subsection{Model Training}
\label{sec:methods_train}
There are three targets~\cref{eq: cnp_llh_1},~\cref{eq: tcl_7},~\cref{eq: fcl_9} involved within each episode.
Nonetheless, three objectives have respective goals and different parameters to learn.
With the generative observation reconstruction objective~$\mca{L}_{\mathrm{FRL}}$, we maximize~\cref{eq: cnp_llh_1} \ie the conditional expectation.
In practice, we minimize the negative log-likelihood function by negating~\cref{eq: cnp_predict_4}.
At the same time, two contrastive losses $\mca{L}_{\mathrm{TCL}}$ and $\mca{L}_{\mathrm{FCL}}$ are optimized prior to estimating the conditional expectation.
When adding them to our training objective with $\alpha \geq 0$ for $\mca{L}_{\mathrm{TCL}}$ and $\beta \geq 0$ for $\mca{L}_{\mathrm{FCL}}$ as tradeoff parameters, we summarize the overall objective
\begin{equation}
    \begin{aligned}
        \arg \min_{\Theta} \;    & \mbb{E}_{f \sim F} \; [ \mca{L}_{\mathrm{FRL}_{\phi, \psi, g, h_{C}}}^{f}                                     \\
                                 & - \alpha \mca{L}_{\mathrm{TCL}_{h_{P}, \varphi, \rho_{P}}} - \beta \mca{L}_{\mathrm{FCL}_{h_{F}, \rho_{F}}} ] \\
        \mathrm{where} \; \Theta & = \{ \phi, \psi, g, h_{C}, h_{P}, h_{F}, \rho_{F}, \rho_{P}, \varphi \}
    \end{aligned}
\end{equation}
The procedure of episodic training is shown in~\cref{algo:training}.
Note that forloops in the description is for illustration, we apply batch processing for parallelism in practice.

\section{Empirical Studies}
\label{sec:experiments}
We empirically validate the proposed CCNP on a broad range of time-series data encompassing 1D, 2D function regression and multi-object trajectory prediction.
Our main interests fall onto the following three questions:
\begin{itemize}
    \item[i.] Do introduce auxiliary contrastive losses help to increase the predictive performance of CNP?
    \item[ii.] If so, does explicitly model {\tt FCL} beneficial for model adaptation?
    \item[iii.] Is CCNP capable of handling against high-dimensional observations than CNP?
\end{itemize}
We attempt to address these research questions by reporting the quantitative experimental results and discussing associated explanations.
In~\cref{sec:experiments_performance} we compare the predictive regression performance on three scopes of datasets to answer question i).
Then, we discuss if the meta-transferability can be improved by {\tt FCL} in~\cref{sec:experiments_meta}.
Moreover, in~\cref{sec:experiments_scalability} we look into if the introduced {\tt TCL} makes CNP scalable to higher-dimensional temporal sequences.
\subsection{Experiments Setup}
\label{sec:experiments_setup}
\noindent {\bf Datasets}.
We briefly describe the datasets covered in this work. See Appendix A.1 for detailed dataset description.

For {\it 1D Functions}, we practice few-shot regression over four different function families, including sinusoids, exponentials, damped oscillators and straight lines.
Each function family is depicted with the linear combination of a amplitude $\alpha \sim \mca{U}(\alpha_{\mathrm{floor}}, \alpha_{\mathrm{ceil}})$ and a phase $\beta \sim \mca{U}(\beta_{\mathrm{floor}}, \beta_{\mathrm{floor}})$ in a closed-form.
For each function family, we construct a meta-dataset by randomly sampling different $\alpha$ and $\beta$ for each instantiation, with each dataset is composed of 490 training, 10 validation and 10 test sequences.
In evaluation, the models are tested 5-shot, 10-shot and 20-shot setting (\ie the size of context set $|D_{C}|$ ), respectively.
All models are trained for 25 episodes.

For {\it 2D Population dynamics}, we study a predator-prey system, where the dynamics can be modeled by the Lotka-Volterra (LV) Equations~\cite{wilkinson2018stochastic}.
The populations of predator $y_{1}$ and prey $y_{2}$ are mutually influenced after each time increment, with the interactions between two species being dominated by the greeks coefficients $\alpha, \beta, \delta, \gamma$.
Expressly, $\alpha$ and $\gamma$ represent predator and prey's birth rates, while the $\beta$ and $\delta$ are associated with the death rates.
We conducted the experiments with two sets of system configurations.
For one setting where initial populations $y_{1} \in [50, 100)$ and $y_{2} \in [100, 150)$ are provided, we fix a set of greeks coefficients $(\alpha, \beta, \delta, \gamma) = (\frac{2}{3}, \frac{4}{3}, 1, 1)$.
In another setting, the initial populations are fixed to 160 and 80, while the greeks are randomly drawn from the predefined range to form each function instantiation.
For both configurations, we generate 180 training, 10 validation and 10 test sequences with 100 times evolvement in each instantiation.
We run 200 episodes for model training, where the sizes of context set and extra target set within each episode are randomly drawn from $[1, 80]$ and $[0, 20]$ therein.

\begin{table*}
    \resizebox{\linewidth}{!}{%
        \begin{tabular}{@{}l|ccc|ccc|ccc|ccc@{}}
            \toprule
            {\bf Model / Data} & \multicolumn{3}{c}{{\bf Sinusoids}}                                                            & \multicolumn{3}{c}{{\bf Exponential}} & \multicolumn{3}{c}{{\bf Linear}} & \multicolumn{3}{c}{{\bf Oscillators}}                                                                                                                                                                                                                                                                 \\
            \midrule
                               & \multicolumn{12}{|c}{{\bf Predictive Log-likelihood ($\times 10^{-2}$)} $\; \uparrow\uparrow$}                                                                                                                                                                                                                                                                                                                                                                                    \\
            \midrule
                               & {\bf 5-shot}                                                                                   & {\bf 10-shot}                         & {\bf 20-shot}                    & {\bf 5-shot}                          & {\bf 10-shot}                 & {\bf 20-shot}                 & {\bf 5-shot}                  & {\bf 10-shot}                 & {\bf 20-shot}                 & {\bf 5-shot}                  & {\bf 10-shot}                 & {\bf 20-shot}                 \\
            {\bf CNP}          & 0.872 $\pm$ 0.292                                                                              & \underline{1.024 $\pm$ 0.117}         & 1.084 $\pm$ 0.061                & 1.298 $\pm$ 0.028                     & 1.313 $\pm$ 0.036             & 1.335 $\pm$ 0.019             & 0.393 $\pm$ 0.140             & 0.428 $\pm$ 0.172             & 0.531 $\pm$ 0.070             & 1.188 $\pm$ 0.083             & 1.233 $\pm$ 0.031             & 1.243 $\pm$ 0.051             \\
            {\bf AttnCNP}      & \underline{0.902 $\pm$ 0.228}                                                                  & 1.004 $\pm$ 0.143                     & 1.079 $\pm$ 0.120                & 1.343 $\pm$ 0.024                     & 1.353 $\pm$ 0.010             & 1.361 $\pm$ 0.011             & \underline{0.545 $\pm$ 0.141} & \underline{0.620 $\pm$ 0.250} & \underline{0.706 $\pm$ 0.198} & 1.242 $\pm$ 0.064             & 1.285 $\pm$ 0.039             & 1.312 $\pm$ 0.026             \\
            {\bf ConvCNP}      & 0.370 $\pm$ 0.326                                                                              & 0.694 $\pm$ 0.281                     & {\bf 1.331 $\pm$ 0.226}          & {\bf 1.807 $\pm$ 0.310}               & {\bf 2.069 $\pm$ 0.234}       & {\bf 2.488 $\pm$ 0.294}       & -0.646 $\pm$ 0.331            & -0.348 $\pm$ 0.534            & 0.160 $\pm$ 0.342             & {\bf 1.345 $\pm$ 0.101}       & {\bf 1.625 $\pm$ 0.193}       & {\bf 2.113 $\pm$ 0.313}       \\
            {\bf CCNP} (Ours)  & {\bf 1.073 $\pm$ 0.086}                                                                        & {\bf 1.144 $\pm$ 0.137 }              & \underline{1.240 $\pm$ 0.067}    & \underline{1.356 $\pm$ 0.013}         & \underline{1.364 $\pm$ 0.020} & \underline{1.372 $\pm$ 0.007} & {\bf 1.021 $\pm$ 0.120}       & {\bf 1.130 $\pm$ 0.075}       & {\bf 1.178 $\pm$ 0.123}       & \underline{1.344 $\pm$ 0.014} & \underline{1.364 $\pm$ 0.009} & \underline{1.372 $\pm$ 0.002} \\
            \midrule
                               & \multicolumn{12}{|c}{{\bf Reconstruction Error ($\times 10^{2}$)} $\; \downarrow\downarrow$}                                                                                                                                                                                                                                                                                                                                                                                      \\
            \midrule
            {\bf CNP}          & 1.135 $\pm$ 0.803                                                                              & \underline{0.714 $\pm$ 0.255}         & 0.571 $\pm$ 0.089                & 0.164 $\pm$ 0.058                     & 0.129 $\pm$ 0.062             & 0.090 $\pm$ 0.041             & 2.657 $\pm$ 0.760             & 2.726 $\pm$ 0.934             & 2.316 $\pm$ 0.447             & 0.393 $\pm$ 0.182             & 0.291 $\pm$ 0.077             & 0.279 $\pm$ 0.108             \\
            {\bf AttnCNP}      & \underline{1.105 $\pm$ 0.832}                                                                  & 0.784 $\pm$ 0.400                     & 0.549 $\pm$ 0.182                & \underline{0.075 $\pm$ 0.056}         & \underline{0.055 $\pm$ 0.025} & \underline{0.039 $\pm$ 0.021} & \underline{1.688 $\pm$ 0.57}  & \underline{1.915 $\pm$ 0.836} & \underline{1.357 $\pm$ 0.697} & \underline{0.25 $\pm$ 0.12}   & \underline{0.169 $\pm$ 0.083} & 0.127 $\pm$ 0.047             \\
            {\bf ConvCNP}      & 3.788 $\pm$ 2.222                                                                              & 1.490 $\pm$ 1.213                     & \underline{0.353 $\pm$ 0.183}    & 1.289 $\pm$ 0.731                     & 0.541 $\pm$ 0.347             & 0.132 $\pm$ 0.072             & 48.367 $\pm$ 59.953           & 21.610 $\pm$ 24.210           & 6.777 $\pm$ 7.635             & 0.897 $\pm$ 0.426             & 0.362 $\pm$ 0.165             & \underline{0.091 $\pm$ 0.045} \\
            {\bf CCNP} (Ours)  & {\bf 0.479 $\pm$ 0.231}                                                                        & {\bf 0.420 $\pm$ 0.312}               & {\bf 0.230 $\pm$ 0.112}          & {\bf 0.041 $\pm$ 0.029}               & {\bf 0.033 $\pm$ 0.041}       & {\bf 0.014 $\pm$ 0.009}       & {\bf 0.519 $\pm$ 0.260}       & {\bf 0.333 $\pm$ 0.097}       & {\bf 0.277 $\pm$ 0.190}       & {\bf 0.068 $\pm$ 0.028}       & {\bf 0.036 $\pm$ 0.017}       & {\bf 0.019 $\pm$ 0.005}       \\
            \bottomrule
        \end{tabular}%
    }
    \caption{Comparison between CCNP and the baseline CNP models. The upper half shows the predictive log-likelihood ($\uparrow\uparrow$ the higher the better) of each model on the target set for each function family. The lower half shows reconstruction error ($\downarrow\downarrow$ the lower the better) on the target indices. Results are reported with $\pm$ 1 standard deviation over 6 different random seeds. Results in {\bf Bold} indicate the best performances, and \underline{underlined} refer to the second-best.}
    \label{tab:1d_results}
\end{table*}

For {\it Higher-dimensional sequence prediction}, we study a synthetic bouncing ball system and rotating MNIST data\footnote{https://github.com/cagatayyildiz/ODE2VAE}, derived from generative temporal modeling scenarios.
In the bouncing ball system, each trajectory (samples sequence) contains the movements of three interacting balls within a rectangular box.
Each timestep is framed as a $32 * 32$ image.
The models are tasked with inferring the locations of the balls as well as interaction rules between them, and reconstructing the pixels in unobserved timesteps, without prior assumption on details of the scene (e.g., ball count and velocities).
We randomly grab 10000 and 500 trajectories for training and testing, where each trajectory consists of 20 simulated time steps of motion.
The number of context samples and extra target samples are fixed to 5 and 10.
In RotMNIST data, we predict 16 rotation angles of handwritten $3$ digits in the shape of $28 * 28$ grayscale pixels sequences.
We follow the same settings as in~\cite{yildiz2019ode2vae} where the 400 sequences with a 9:1 ratio of train-test split to run the experiments.
Similarly, in each sequence, the model learns from $5$ context samples randomly drawn from $16$ frames is tasked with predicting the remaining frames.

\

\noindent {\bf Baselines}.
Resembling most of related studies, we consider vanilla CNP (CNP)~\cite{garnelo2018conditional}, Attentive CNP (AttnCNP)~\cite{kimanp18}, and Convolutional CNP (ConvCNP)~\cite{Gordon2019ConvCNP} as baselines.

\

\noindent {\bf Evaluation Metrics}.
To answer question i), we evaluate the models' performances in terms of predictive mean log-likelihood and reconstruction error measured by mean squared error (MSE) as usual for CNPs.
In~\cref{sec:experiments_meta} we additionally validate the transferability of meta-representation by predicting the function coefficients of a specific data-generating function.
MSE is adopted as the measurement.

\subsection{Exact Reconstruction Results}
\label{sec:experiments_performance}
\begin{figure}
    \centering
    \includegraphics[width=0.98\linewidth]{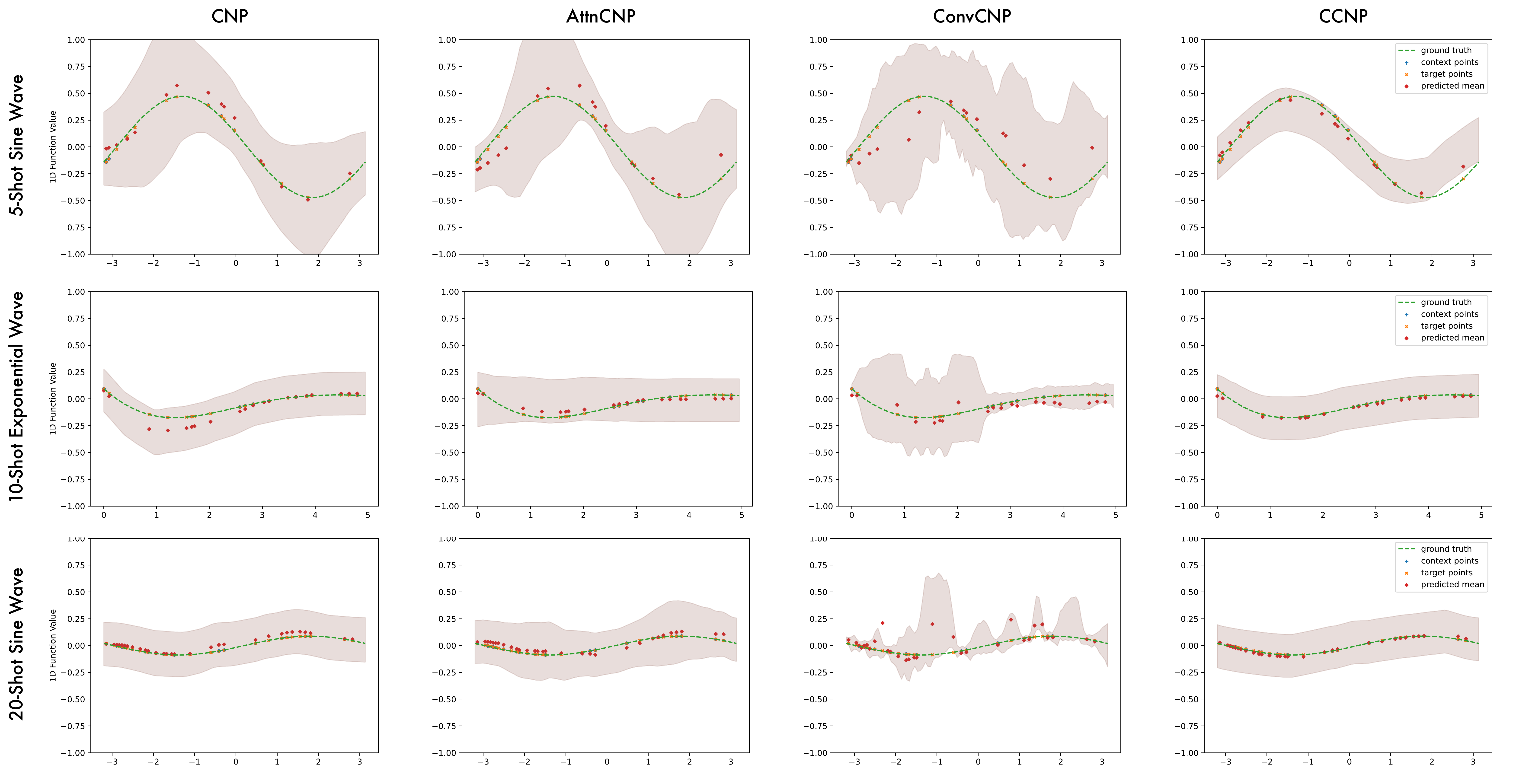}
    \label{fig: 1d_pred}
    \caption{The qualitative comparison of CCNP and CNPs baselines on three different function families with 5-shot, 10-shot and 20-shot regression. The shown results in each column are model predictions on the shared target samples, where the plottings from left to right are of CNP, AttnCNP, ConvCNP and CCNP therein, respectively. Dashed line is the ground-truth function value, with red diamonds representing the predictive mean values $\bs{\mu}$ and the shaded regions indicating $95\%$ confidence interval ($\bs{\mu} \pm 2\bs{\sigma}$).}
\end{figure}
\noindent {\bf 1D Few-Shot Function Regression}.
\cref{tab:1d_results} summarizes our finding in running few-shot function regression over four groups of data-generating functions.
In this task, the predictive model takes a context set $D_{C}^{f}$ of an instantiation $f$.
The contextual information is extracted to a global representation $\repf{}{f}$, which is used then to produce the Gaussian mean of $\bs{y}_{t}^{f}$ and the predictive variance in the target index $x_{t}$, with $x_{t} \in D_{T}^{f}$.
\cref{fig: 1d_pred} illustrates that CCNP can make reasonable predictions with higher confidence (lower variance) even only 5 context points are provided.
CCNP remarkably outperforms all the competitive baselines in relating to both predictive log-likelihood on context sets and reconstruction error on target sets.

\noindent {\bf 2D Population Dynamics Prediction}.
We also compare the function reconstruction capabilities on the Lotka-Volterra system of CNP, AttnCNP and CCNP.
On a par with 1D tasks, we report the predictive log-likelihood of target indices on the validation set.
Under both configurations, three models can converge within 100 epochs, whilst CCNP can fit LV equations with fewer epochs when the initial populations are provided.
For qualitative comparisons, the maximum likelihood of CNP reaches $1.626 \pm 0.24$ and $1.652 \pm 0.13$ under two modes.
AttnCNP performs faster adaptation in the former case, which, however is not compared favorably with the ones achieved by CCNP after ~100 epochs, indicated by $2.059 \pm 0.11 (\mathrm{AttnCNP})  < 3.794 \pm 0.58 (\mathrm{CCNP}) $).
\begin{figure}
    \centering
    \begin{subfigure}[b]{0.45\linewidth}
        \centering
        \includegraphics[width=\linewidth]{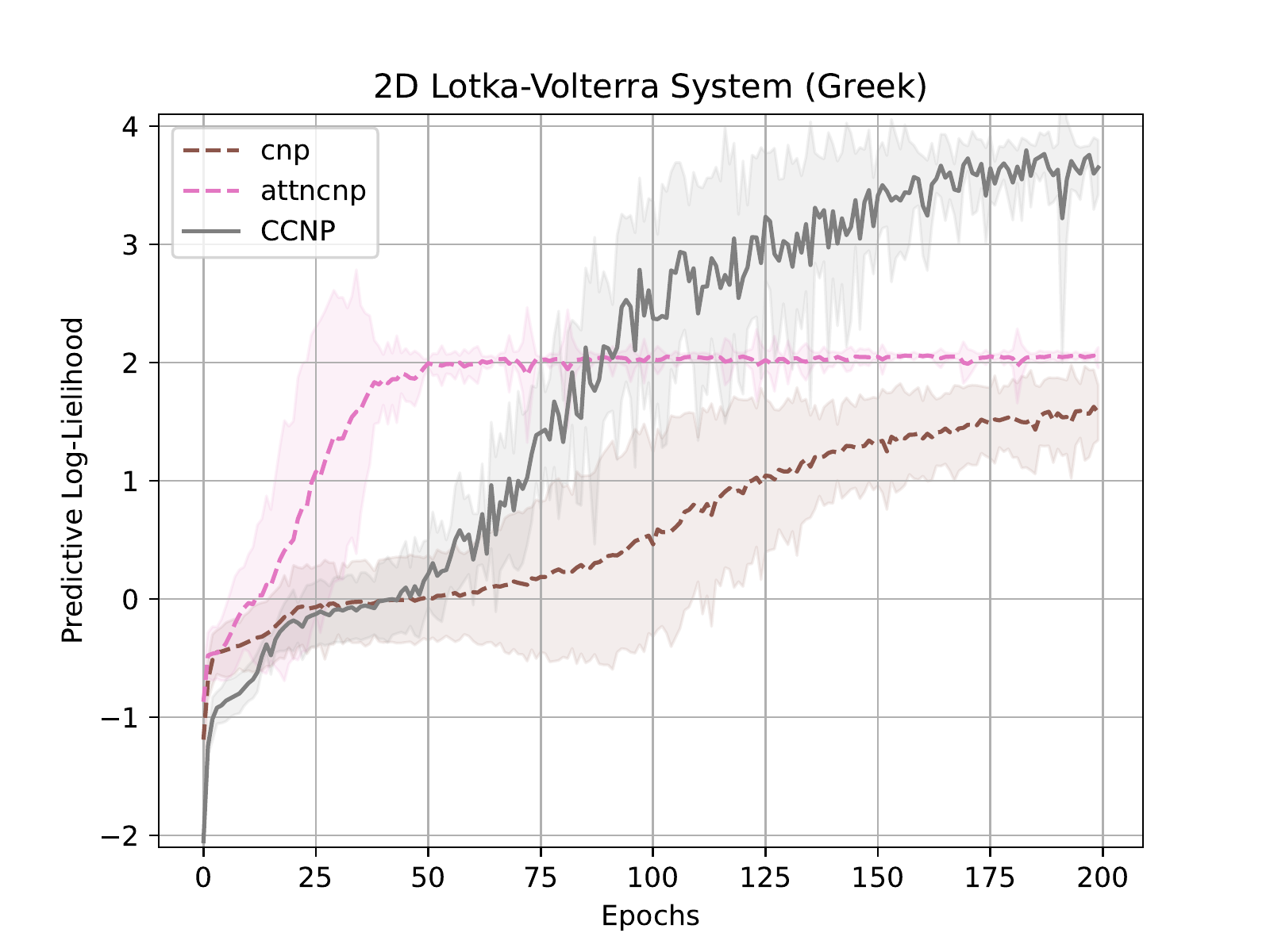}
        \caption{Log-likelihood with random initial populations.}
        \label{fig:4a_lv_greek}
    \end{subfigure}
    \begin{subfigure}[b]{0.45\linewidth}
        \centering
        \includegraphics[width=\linewidth]{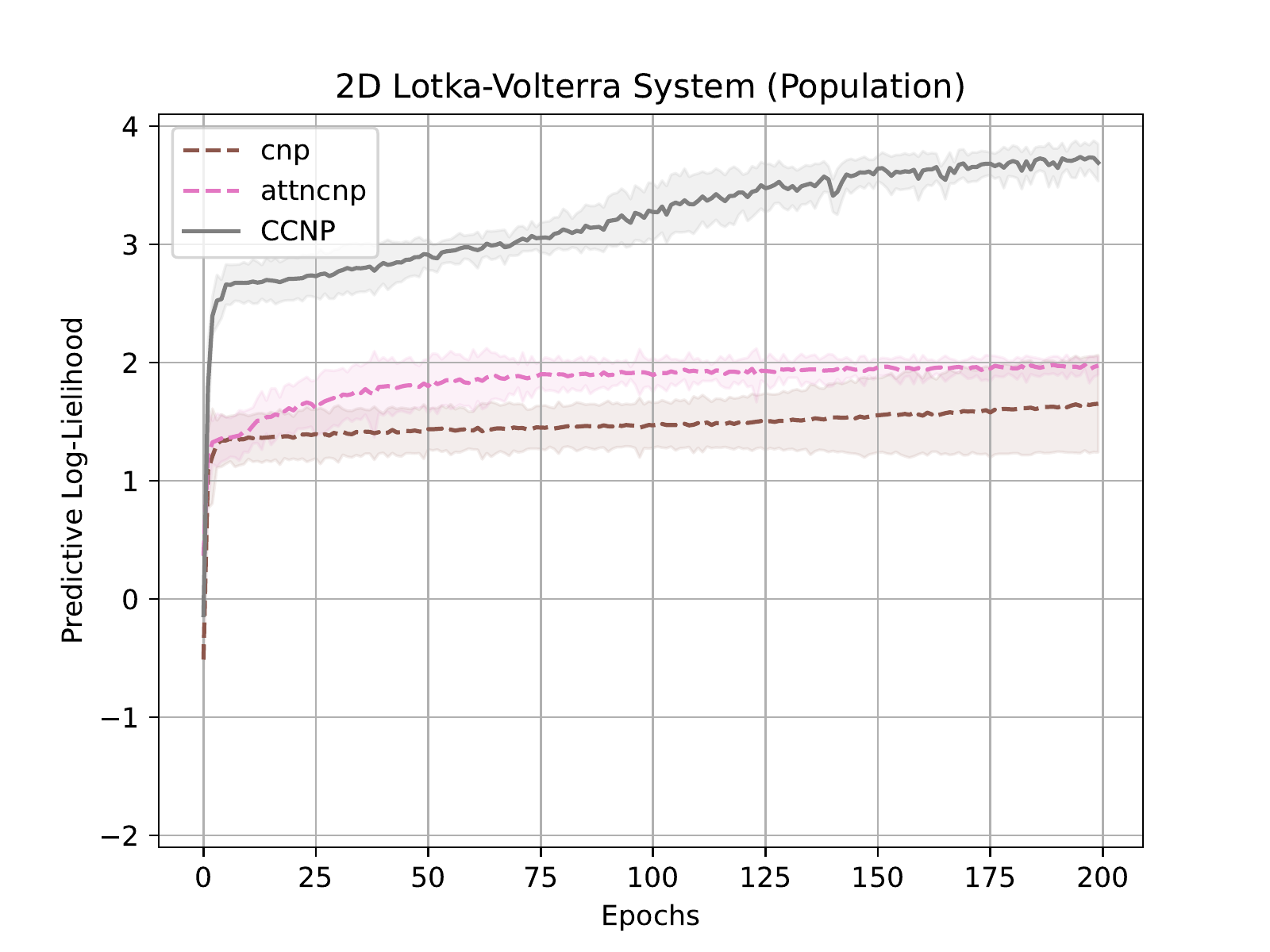}
        \caption{Log-likelihood with fixed initial populations.}
        \label{fig:5b_lv_population}
    \end{subfigure}
    \caption{Predictive performances of 2D predator-prey dynamics. Left reveals the simulation with randomly initialized population and fixed greeks coefficients, and vice versa in the right. Grey solid line represents the predictive mean of CCNP over 6 seeds. Pink and brown dashed lines correspond to CNP and AttnCNP, respectively. We denote $ 95\% $ confidence interval with shaded areas.}
\end{figure}

\noindent {\bf Higher-dimensional Sequence Prediction}.
We provide both qualitative and quantitative comparisons among CNP, AttnCNP and CCNP for predicting the evolvement of higher-dimensional image sequences (see~\cref{fig:5c_image_comparisons} and~\cref{tab:ablation}).
With the same encoding structure and training condition, CCNP returns lower MSE in reconstruction error across multiple prediction steps on both datasets.
This can also be verified in quantitative results, where the second row consists of predictive reconstruction of CNP, while CCNP is reported within the third row.
While CNP produces blur results at some query indices, CCNP can reconstruct more accurate values more confidently.
For both CNP and CCNP, there is no significant performance drop when prediction sizes are extended by 3-4 times longer, reflecting that CNPs have a great potential in handling low data problems.

We further provide experiments in terms of additional 1D datasets and running efficiency (See Appendix B.).

\begin{figure*}
    \centering
    \includegraphics[width=0.9\textwidth]{figures/6_imgseq.pdf}
    \caption{The qualitative and quantitative comparisons of CNP and CCNP on two image sequences datasets.}
    \label{fig:5c_image_comparisons}
\end{figure*}

\subsection{Transferability of Meta-representation}
\label{sec:experiments_meta}
\begin{table}
    \centering
    \resizebox{0.95\linewidth}{!}{%
        \begin{tabular}{@{}c|cc|cc@{}}
            \toprule
            \multirow{2}{*}{{\bf Model / Data}} & \multicolumn{2}{c}{{\bf RotMNIST}} & \multicolumn{2}{|c}{{\bf BouncingBall}}                                                                   \\
                                                & 10                                 & 16                                      & 10                             & 20                             \\
            \midrule
            CNP                                 & 0.0132 $\pm$ 0.002                 & 0.0133 $\pm$ 0.002                      & 0.0579 $\pm$ 0.004             & 0.0607 $\pm$ 0.004             \\
            CCNP~(-Attn)                        & \underline{0.0085 $\pm$ 0.003}     & \underline{0.0087 $\pm$ 0.001}          & \underline{0.0465 $\pm$ 0.003} & \underline{0.0490 $\pm$ 0.003} \\
            CCNP~(-TCL)                         & 0.0103 $\pm$ 0.001                 & 0.0103 $\pm$ 0.001                      & 0.0505 $\pm$ 0.003             & 0.0526 $\pm$ 0.003             \\
            \midrule
            CCNP~(Full)                         & {\bf 0.0066 $\pm$ 0.001}           & {\bf 0.0068 $\pm$ 0.001}                & {\bf 0.0458 $\pm$ 0.004}       & {\bf 0.0489 $\pm$ 0.004}       \\
            \bottomrule
        \end{tabular}%
    }
    \caption{Prediction performance on high-dimensional images sequences with {\bf ablation studies}. We report MSE~$(\downarrow \downarrow)$ obtained on RotMNIST and BouncingBall data, whereby we also conduct ablation studies for removing a) position-aware aggregation (indicated by CCNP~(-Attn)); b) {\tt TCL} branch (CCNP~(-TCL)). Results in {\bf Bold} indicate the best performances and \underline{underlined} refer to the second-best, with $\pm$ 1 standard deviation over 6 runs for each.}
    \label{tab:ablation}
\end{table}

Besides the evaluation of reconstructing observations, we are also interested in how well the model can recognize the particular generating function, which is associated with the transferability of meta-representation~\cite{gondal2021function}.
We study the following experiment \ie function coefficient inference in closed-formed 1D sinusoid functions to discuss question ii).

\begin{figure}
    \centering
    \begin{subfigure}[b]{0.5\linewidth}
        \centering
        \includegraphics[width=\linewidth]{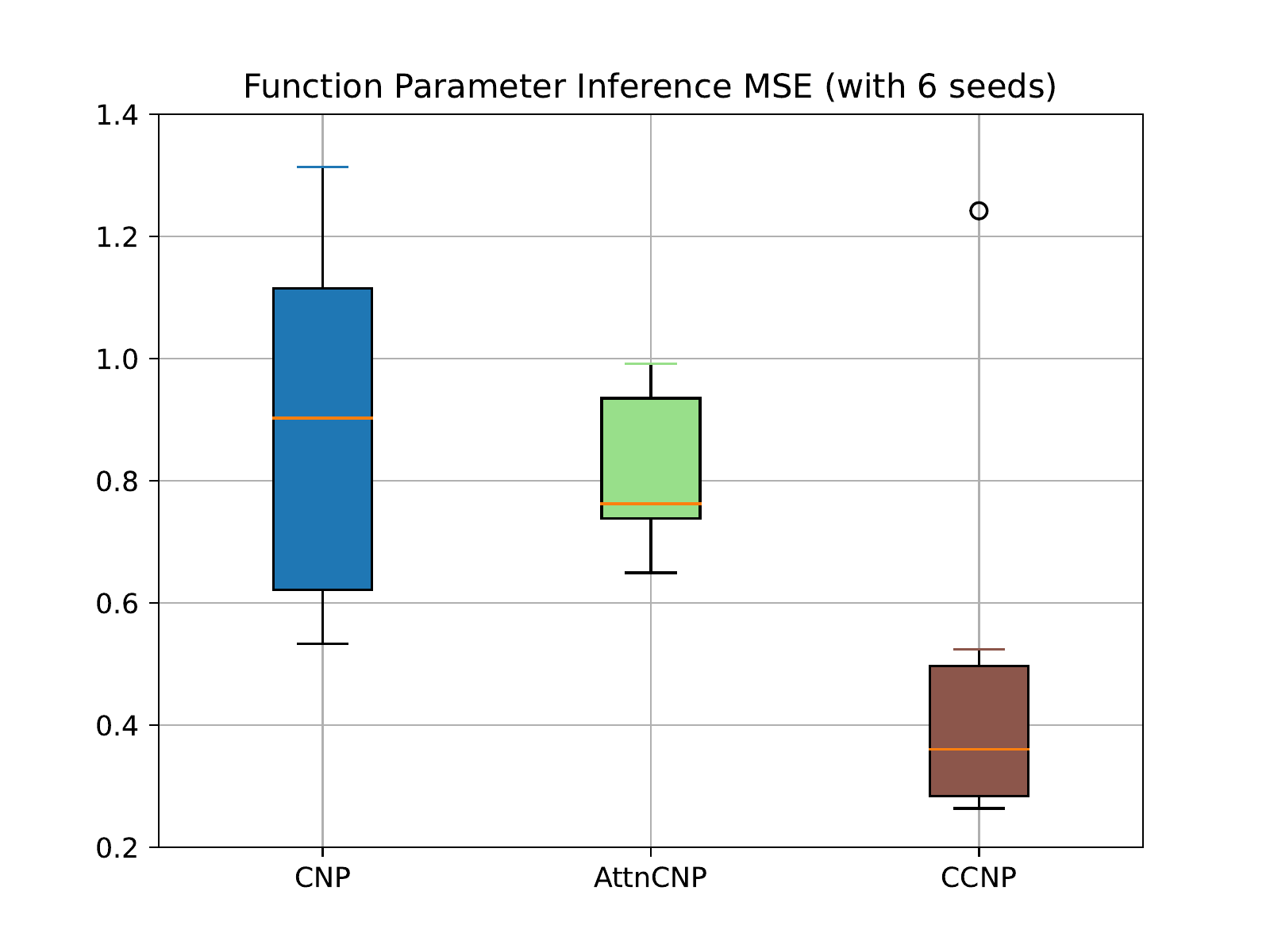}
        \caption{Results of function coefficient inference over 6 different seeds with blue, green and brown box referring to CNP, AttnCNP and CCNP, respectively.}
        \label{fig:5a_param_result}
    \end{subfigure}
    \begin{subfigure}[b]{0.49\linewidth}
        \centering
        \includegraphics[width=\linewidth]{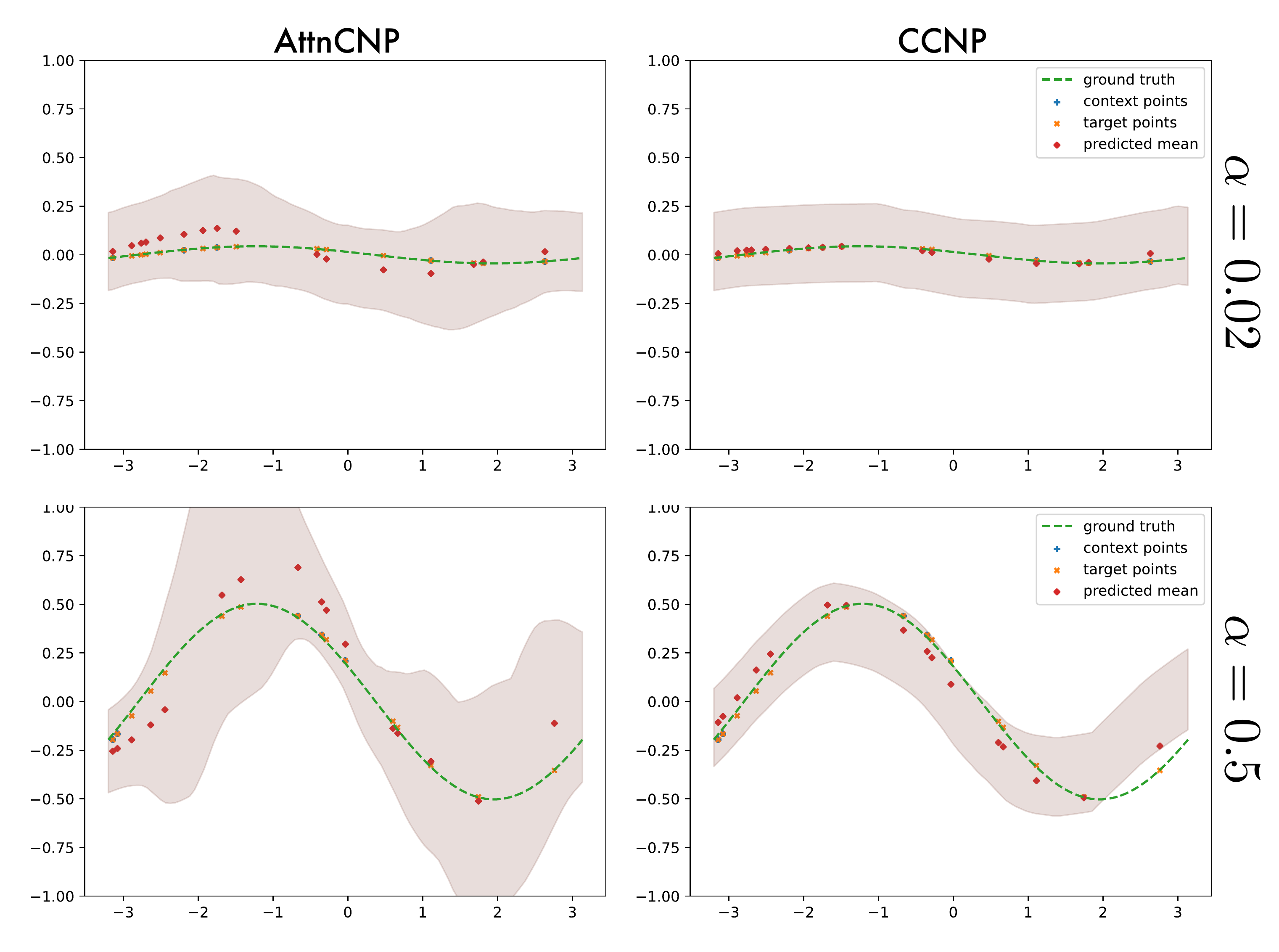}
        \caption{Example of identifying shifted function coefficient amplitude $\alpha$. The right column (CCNP) adapts to different instantiations, where the left (AttnCNP) fails to do so.}
        \label{fig:5b_amplitude_shift}
    \end{subfigure}
    \caption{Comparisons of CNP, AttnCNP and CCNP on the transferability of meta-representations.}
\end{figure}

The goal of this task is to predict the amplitude $\alpha$ and phase $\beta$ of the sinusoid function $f$, given a randomly sampled context set $D_{C}^{f}$.
For each model, we fix the parameters of contextual representation $\repf{}{f}$ pre-trained on function regression tasks, of which we append a classifier on top to perform coefficient inference.
\cref{fig:5a_param_result} describes performances of CNP, AttnCNP and CCNP on this task, where CCNP produces the lowest MSE in evaluating the prediction of $\alpha$ and $\beta$, with ~-0.5 (to CNP) and ~-0.4 (to AttnCNP) of gap over 6 seeds.
We conclude that CCNP demonstrates higher transferability of meta-representation, benefited by {\tt FCL} regularizing its capability of aligning function instantiations.
For quantitative illustration, we showcase that AttnCNP struggles at adapting to shifted $\alpha$ within the same function family while CCNP maintains decent capability~(\cref{fig:5b_amplitude_shift}).

\subsection{Scalability to Higher-Dimensional Data}
\label{sec:experiments_scalability}
We now investigate question iii) \ie whether {\tt TCL} is beneficial to the scalability of CNPs.
The results in~\cref{tab:ablation} suggest the prediction performances given by different design choices of CCNP.
Removing either encoding self-attention or {\tt TCL} leads to poor performance in predicting high-dimensional data.
Stripping TCL from CCNP brings in decreased results in predicting both datasets, which can be attributed to additional efficiency provided by contrastive loss against high-dimensional data.
The position-aware encoding also boosts the performance of {\tt TCL} as the combination of the two is better than if either were applied alone.

\section{Conclusion}
We present Contrastive Conditional Neural Processes, a hybrid generative-contrastive model that explores the complementary advantages of generative and contrastive approaches in meta-learning, towards scaling CNPs to handle high-dimensional and noisy time-series.
Generative CNPs are augmented by two contrastive, hierarchically organized branches.
While the in-instantiation temporal contrastive branch revolves around extracting high-level abstraction of observations, the cross-instantiation function contrastive branch cope with supervision-collapse of common generative meta-learning models.
Improved reconstruction performances on diverse time-series datasets demonstrate that, adding contrastive regularization to CNPs not only directly facilitates discriminative down-stream tasks, but also contributes significantly to generative reconstruction tasks.

\noindent {\bf Limitation}.
In this work, we focus on improving the efficacy of CNPs with two auxiliary contrastive regularizers.
The costs associated with constructing tuples and comparing pairwise similarity are, however, causing a loss in computation efficiency.
Such tradeoffs may be addressed by exploring more efficient contrastive sampling methods.

    {\small
        \bibliographystyle{ieee_fullname}
        \bibliography{egbib}

\begin{thebibliography}{10}\itemsep=-1pt

\bibitem{chen2020simple}
Ting Chen, Simon Kornblith, Mohammad Norouzi, and Geoffrey Hinton.
\newblock A simple framework for contrastive learning of visual
  representations.
\newblock In {\em International conference on machine learning}, pages
  1597--1607. PMLR, 2020.

\bibitem{chen2017learning}
Yutian Chen, Matthew~W Hoffman, Sergio~G{\'o}mez Colmenarejo, Misha Denil,
  Timothy~P Lillicrap, Matt Botvinick, and Nando Freitas.
\newblock Learning to learn without gradient descent by gradient descent.
\newblock In {\em International Conference on Machine Learning}, pages
  748--756. PMLR, 2017.

\bibitem{doersch2020crosstransformers}
Carl Doersch, Ankush Gupta, and Andrew Zisserman.
\newblock Crosstransformers: spatially-aware few-shot transfer.
\newblock {\em Advances in Neural Information Processing Systems},
  33:21981--21993, 2020.

\bibitem{Foong2020ConvNP}
Andrew Foong, Wessel Bruinsma, Jonathan Gordon, Yann Dubois, James Requeima,
  and Richard Turner.
\newblock Meta-learning stationary stochastic process prediction with
  convolutional neural processes.
\newblock In H. Larochelle, M. Ranzato, R. Hadsell, M.~F. Balcan, and H. Lin,
  editors, {\em Advances in Neural Information Processing Systems}, volume~33,
  pages 8284--8295. Curran Associates, Inc., 2020.

\bibitem{gao2021simcse}
Tianyu Gao, Xingcheng Yao, and Danqi Chen.
\newblock Simcse: Simple contrastive learning of sentence embeddings.
\newblock {\em arXiv preprint arXiv:2104.08821}, 2021.

\bibitem{garnelo2018conditional}
Marta Garnelo, Dan Rosenbaum, Christopher Maddison, Tiago Ramalho, David
  Saxton, Murray Shanahan, Yee~Whye Teh, Danilo Rezende, and SM~Ali Eslami.
\newblock Conditional neural processes.
\newblock In {\em International Conference on Machine Learning}, pages
  1704--1713. PMLR, 2018.

\bibitem{garnelo2018neural}
Marta Garnelo, Jonathan Schwarz, Dan Rosenbaum, Fabio Viola, Danilo~J. Rezende,
  S.~M.~Ali Eslami, and Yee~Whye Teh.
\newblock Neural processes, 2018.

\bibitem{gondal2021function}
Muhammad~Waleed Gondal, Shruti Joshi, Nasim Rahaman, Stefan Bauer, Manuel
  Wuthrich, and Bernhard Sch{\"o}lkopf.
\newblock Function contrastive learning of transferable meta-representations.
\newblock In {\em International Conference on Machine Learning}, pages
  3755--3765. PMLR, 2021.

\bibitem{Gordon2019ConvCNP}
Jonathan Gordon, Wessel~P. Bruinsma, Andrew Y.~K. Foong, James Requeima, Yann
  Dubois, and Richard~E. Turner.
\newblock Convolutional conditional neural processes.
\newblock In {\em 8th International Conference on Learning Representations,
  {ICLR} 2020, Addis Ababa, Ethiopia, April 26-30, 2020}. OpenReview.net, 2020.

\bibitem{gutmann2010noise}
Michael Gutmann and Aapo Hyv{\"a}rinen.
\newblock Noise-contrastive estimation: A new estimation principle for
  unnormalized statistical models.
\newblock In {\em Proceedings of the thirteenth international conference on
  artificial intelligence and statistics}, pages 297--304. JMLR Workshop and
  Conference Proceedings, 2010.

\bibitem{holderrieth2021equivariant}
Peter Holderrieth, Michael~J Hutchinson, and Yee~Whye Teh.
\newblock Equivariant learning of stochastic fields: Gaussian processes and
  steerable conditional neural processes.
\newblock In {\em International Conference on Machine Learning}, pages
  4297--4307. PMLR, 2021.

\bibitem{hospedales2020meta}
Timothy Hospedales, Antreas Antoniou, Paul Micaelli, and Amos Storkey.
\newblock Meta-learning in neural networks: A survey.
\newblock {\em arXiv preprint arXiv:2004.05439}, 2020.

\bibitem{ivanovic2019trajectron}
Boris Ivanovic and Marco Pavone.
\newblock The trajectron: Probabilistic multi-agent trajectory modeling with
  dynamic spatiotemporal graphs.
\newblock In {\em Proceedings of the IEEE/CVF International Conference on
  Computer Vision}, pages 2375--2384, 2019.

\bibitem{kallidromitis2021contrastive}
Konstantinos Kallidromitis, Denis Gudovskiy, Kozuka Kazuki, Ohama Iku, and Luca
  Rigazio.
\newblock Contrastive neural processes for self-supervised learning.
\newblock In {\em Asian Conference on Machine Learning}, pages 594--609. PMLR,
  2021.

\bibitem{KawanoKSIM21}
Makoto Kawano, Wataru Kumagai, Akiyoshi Sannai, Yusuke Iwasawa, and Yutaka
  Matsuo.
\newblock Group equivariant conditional neural processes.
\newblock In {\em 9th International Conference on Learning Representations,
  {ICLR} 2021, Virtual Event, Austria, May 3-7, 2021}. OpenReview.net, 2021.

\bibitem{kimanp18}
Hyunjik Kim, Andriy Mnih, Jonathan Schwarz, Marta Garnelo, Ali Eslami, Dan
  Rosenbaum, Oriol Vinyals, and Yee~Whye Teh.
\newblock Attentive neural processes.
\newblock In {\em Proceedings of the International Conference on Learning
  Representations (ICLR)}, 2019.

\bibitem{DBLP:conf/iclr/KipfPW20}
Thomas~N. Kipf, Elise van~der Pol, and Max Welling.
\newblock Contrastive learning of structured world models.
\newblock In {\em 8th International Conference on Learning Representations,
  {ICLR} 2020, Addis Ababa, Ethiopia, April 26-30, 2020}. OpenReview.net, 2020.

\bibitem{le2018empirical}
Tuan~Anh Le, Hyunjik Kim, Marta Garnelo, Dan Rosenbaum, Jonathan Schwarz, and
  Yee~Whye Teh.
\newblock Empirical evaluation of neural process objectives.
\newblock In {\em NeurIPS workshop on Bayesian Deep Learning}, 2018.

\bibitem{liu2021self}
Xiao Liu, Fanjin Zhang, Zhenyu Hou, Li Mian, Zhaoyu Wang, Jing Zhang, and Jie
  Tang.
\newblock Self-supervised learning: Generative or contrastive.
\newblock {\em IEEE Transactions on Knowledge and Data Engineering}, 2021.

\bibitem{mathieu2021contrastive}
Emile Mathieu, Adam Foster, and Yee Teh.
\newblock On contrastive representations of stochastic processes.
\newblock {\em Advances in Neural Information Processing Systems}, 34, 2021.

\bibitem{messing2009activity}
Ross Messing, Chris Pal, and Henry Kautz.
\newblock Activity recognition using the velocity histories of tracked
  keypoints.
\newblock In {\em 2009 IEEE 12th international conference on computer vision},
  pages 104--111. IEEE, 2009.

\bibitem{pacchiardi2022score}
Lorenzo Pacchiardi and Ritabrata Dutta.
\newblock Score matched neural exponential families for likelihood-free
  inference.
\newblock {\em Journal of Machine Learning Research}, 23(38):1--71, 2022.

\bibitem{Racah2020Slot}
Evan Racah and Sarath Chandar.
\newblock Slot contrastive networks: {A} contrastive approach for representing
  objects.
\newblock {\em CoRR}, abs/2007.09294, 2020.

\bibitem{salzmann2020trajectron++}
Tim Salzmann, Boris Ivanovic, Punarjay Chakravarty, and Marco Pavone.
\newblock Trajectron++: Dynamically-feasible trajectory forecasting with
  heterogeneous data.
\newblock In {\em Computer Vision--ECCV 2020: 16th European Conference,
  Glasgow, UK, August 23--28, 2020, Proceedings, Part XVIII 16}, pages
  683--700. Springer, 2020.

\bibitem{tack2020csi}
Jihoon Tack, Sangwoo Mo, Jongheon Jeong, and Jinwoo Shin.
\newblock Csi: Novelty detection via contrastive learning on distributionally
  shifted instances.
\newblock {\em arXiv preprint arXiv:2007.08176}, 2020.

\bibitem{pmlr-v130-ton21a}
Jean-Francois Ton, Lucian CHAN, Yee Whye~Teh, and Dino Sejdinovic.
\newblock Noise contrastive meta-learning for conditional density estimation
  using kernel mean embeddings.
\newblock In Arindam Banerjee and Kenji Fukumizu, editors, {\em Proceedings of
  The 24th International Conference on Artificial Intelligence and Statistics},
  volume 130 of {\em Proceedings of Machine Learning Research}, pages
  1099--1107. PMLR, 13--15 Apr 2021.

\bibitem{tsai2022conditional}
Yao-Hung~Hubert Tsai, Tianqin Li, Martin~Q. Ma, Han Zhao, Kun Zhang,
  Louis-Philippe Morency, and Ruslan Salakhutdinov.
\newblock Conditional contrastive learning with kernel.
\newblock In {\em International Conference on Learning Representations}, 2022.

\bibitem{oord2019representation}
Aaron van~den Oord, Yazhe Li, and Oriol Vinyals.
\newblock Representation learning with contrastive predictive coding, 2019.

\bibitem{VaswaniAttention17}
Ashish Vaswani, Noam Shazeer, Niki Parmar, Jakob Uszkoreit, Llion Jones,
  Aidan~N. Gomez, Lukasz Kaiser, and Illia Polosukhin.
\newblock Attention is all you need.
\newblock In Isabelle Guyon, Ulrike von Luxburg, Samy Bengio, Hanna~M. Wallach,
  Rob Fergus, S.~V.~N. Vishwanathan, and Roman Garnett, editors, {\em Advances
  in Neural Information Processing Systems 30: Annual Conference on Neural
  Information Processing Systems 2017, December 4-9, 2017, Long Beach, CA,
  {USA}}, pages 5998--6008, 2017.

\bibitem{vilalta2002perspective}
Ricardo Vilalta and Youssef Drissi.
\newblock A perspective view and survey of meta-learning.
\newblock {\em Artificial intelligence review}, 18(2):77--95, 2002.

\bibitem{wilkinson2018stochastic}
Darren~J Wilkinson.
\newblock {\em Stochastic modelling for systems biology}.
\newblock Chapman and Hall/CRC, 2018.

\bibitem{winkens2020contrastive}
Jim Winkens, Rudy Bunel, Abhijit~Guha Roy, Robert Stanforth, Vivek Natarajan,
  Joseph~R Ledsam, Patricia MacWilliams, Pushmeet Kohli, Alan Karthikesalingam,
  Simon Kohl, et~al.
\newblock Contrastive training for improved out-of-distribution detection.
\newblock {\em arXiv preprint arXiv:2007.05566}, 2020.

\bibitem{yildiz2019ode2vae}
Cagatay Yildiz, Markus Heinonen, and Harri Lahdesmaki.
\newblock Ode2vae: Deep generative second order odes with bayesian neural
  networks.
\newblock {\em Advances in Neural Information Processing Systems},
  32:13412--13421, 2019.

\bibitem{you2020graph}
Yuning You, Tianlong Chen, Yongduo Sui, Ting Chen, Zhangyang Wang, and Yang
  Shen.
\newblock Graph contrastive learning with augmentations.
\newblock {\em Advances in Neural Information Processing Systems},
  33:5812--5823, 2020.

\bibitem{ZaheerKRPSS17}
Manzil Zaheer, Satwik Kottur, Siamak Ravanbakhsh, Barnab{\'{a}}s P{\'{o}}czos,
  Ruslan Salakhutdinov, and Alexander~J. Smola.
\newblock Deep sets.
\newblock In Isabelle Guyon, Ulrike von Luxburg, Samy Bengio, Hanna~M. Wallach,
  Rob Fergus, S.~V.~N. Vishwanathan, and Roman Garnett, editors, {\em Advances
  in Neural Information Processing Systems 30: Annual Conference on Neural
  Information Processing Systems 2017, December 4-9, 2017, Long Beach, CA,
  {USA}}, pages 3391--3401, 2017.

\end{thebibliography}
    }

\end{document}


\title{Appendices to Contrastive Conditional Neural Processes}  

\maketitle
\thispagestyle{empty}
\appendix

This contains three sections for describing experimental datasets, additional experimental results and detailed model implementation.

\section{Dataset Description}
\label{sec: dataset}
The experiments in this work are three folds, including 1D time-series function, 2D predator-prey dynamics and high-dimensional time-series datasets.

\subsection{1D time-series functions}
We run few-shot regression on four function families, see~\cref{tab:1d_dataset} for the specific definition of each.
Each function instantiation used in training/validation/testing is generated with respect to the specified range therein.
By following conventional meta-learning setting, instantiations within a batch are randomly sampled from the dataset of sinusoid functions without replacement, \ie $ f_{i} \neq f_{i^{\prime}}, \; \forall f_{i} \in f_{B} $, with $f_{B} = \{ f_{i} \}_{i=1:|B|}$.

We generate 500 samples for each function family with splitting training/validation/testing in the $9:1:1$ ratio, without function overlap.
For each instantiation, context size is sampled from $(0, N]$ ($N$ refers to $N$-shot) with extra target size is drawn from $(0, 10]$ in the training phase, whilst in validation and testing phase the context size is fixed at $N$ and prediction is evaluated on the whole sequence.
The reported results are acquired on the test set.
The dimension of observation space is $\mathcal{Y} \subseteq \mathbb{R}$.

\begin{table}[b]
    \resizebox{\linewidth}{!}{%
        \begin{tabular}{@{}lcccc@{}}
            \toprule
            \multicolumn{5}{c}{{\bf 1D Synthetic Function}}                                                              \\
            \midrule
            {\bf Family}   & {\bf Form}                                  & $\bs{\alpha}$ & $\bs{\beta}$  & $\bs{x}$      \\
            \midrule
            Sinusoid       & $y = \alpha \sin (x - \beta)$               & $(-1, 1)$     & $(-0.5, 0.5)$ & $(-\pi, \pi)$ \\
            Exponentials   & $y = \alpha \times \exp (x - \beta) $       & $(-1, 1)$     & $(-0.5, 0.5)$ & $(-1, 4)$     \\
            Oscillators    & $y = \alpha \sin (x - \beta) \exp (-0.5 t)$ & $(-1, 1)$     & $(-0.5, 0.5)$ & $(0, 5)$      \\
            Straight lines & $y = \alpha \, x + \beta $                  & $(-1, 1)$     & $(-0.5, 0.5)$ & $(0, 5)$      \\
            \bottomrule
        \end{tabular}
    }
    \caption{Details of 1D times-series functions. Columns of $\alpha, \beta, x$ corresponds to the range where a function instantiation is randomly sampled therein (e.g., $f_{1} = -0.5 \sin ( x - 0.3)$ and $f_{2} = 0.4 \sin (x + 0.1)$). }
    \label{tab:1d_dataset}
\end{table}

\subsection{1D GP-generated functions}
In addition to the above four function families, we also conduct 1D regression experiments where the data-generating functions are Gaussian Processes with different kernels.
We use the kernels as in AttnCNP and ConvCNP, including RBF, Periodic and Noisy Mat\'ern:
\begin{equation}
    \begin{aligned}
        k(x_{i}, x_{j})_{\mathrm{RBF}} & = \exp \left( - \frac{d(x_i, x_j)^2}{2 l^{2}}\right)                                                  \\
        k(x_{i}, x_{j})_{\mathrm{PER}} & = \exp \left(-\frac{2 \sin ^{2} \left(\pi d\left(x_{i}, x_{j}\right) / p \right)}{l^{2}}\right)       \\
        k(x_{i}, x_{j})_{\mathrm{MAT}} & = \frac{1}{\Gamma(\nu) 2^{\nu-1}}\left(\frac{\sqrt{2 \nu}}{l} d\left(x_{i}, x_{j}\right)\right)^{\nu} \\
                                       & \quad K_{\nu}\left(\frac{\sqrt{2 \nu}}{l} d\left(x_{i}, x_{j}\right)\right) + \epsilon                \\
    \end{aligned}
\end{equation}
The number of instantiations we generate for each kernel is 4096, of which 256 are used for testing and 256 for validation, respectively.
Other than that, we use the same sampling strategy as time-series functions, and perform $N$-shot regression.

\subsection{2D predator-prey dynamics}
To model the 2D population dynamics, we fit the Lotka-Volterra equations~(LV) with CNPs.
Given values of $y_1$ and $y_2$ at initial time index $x=0$, the poluations of both species vary after each time increment based on interaction coefficients $\alpha, \beta, \delta, \gamma$.
The growth rates are defined by
\begin{equation}
    \begin{aligned}
        \nabla_{x}y_{1} & = \alpha y_{1} - \beta y_{1} y_{2}  \\
        \nabla_{x}y_{2} & = \delta y_{1} y_{2} - \gamma y_{2}
    \end{aligned}
\end{equation}
with respect to time increment $x: 0 \rightarrow x_{\mathrm{max}}$.
Thus, in each simulated trajectory, the populations at each time index are determined by the initial values of $y_{1}, y_{2}, \alpha, \beta, \gamma, \delta$.
We consider two modes of simulation.
One is {\it Greek} mode where $\alpha, \beta, \gamma, \delta$ are set to fixed values with $y_{1}$ and $y_{2}$ are randomly sampled from the given range, while in {\it Population} mode $y_{1}$ and $y_{2}$ are assigned with fixed initial numbers with $\alpha, \beta, \gamma, \delta$ become random variables~(\cref{tab:2d_dataset}).
For both modes, we run 200 trials as the meta-dataset with accumulating 150 timesteps, \ie $x_{\mathrm{max}}=150$ for every trial.
We generate 200 samples for each function family with splitting training/validation/testing in the $9:1:1$ ratio, without function overlap.
For each trajectory, context size is sampled from $(0, 80]$ with extra target size is drawn from $(0, 20]$ in the training phase, whilst in validation and testing phase the context size is fixed at 80 and prediction is evaluated on the whole trajectory.
The reported results are acquired on the test set.
In this case, the dimension of observation space is $\mathcal{Y} \subseteq \mathbb{R}^{2}$.

\begin{table}
    \resizebox{\linewidth}{!}{%
        \begin{tabular}{@{}lcccccc@{}}
            \toprule
            \multicolumn{7}{c}{ {\bf 2D Population Dynamics} }                                                  \\
            \midrule
            {\bf Mode} & $y_{1}$    & $y_{2}$    & $\bs{\alpha}$ & $\bs{\beta}$ & $\bs{\gamma}$ & $\bs{\delta}$ \\
            \midrule
            Greek      & (0.5, 2.0) & (0.5, 2.0) & $4/3$         & $2/3$        & $1$           & $1$           \\
            Population & 1.6        & 0.8        & (0.9, 1.1)    & (0.05, 0.15) & (1.25, 1.75)  & (0.5, 1.0)    \\
            \bottomrule
        \end{tabular}
    }
    \caption{Details of 2D LV systems. $y_1$ and $y_2$ denote the initial population of predator and prey, resepctively. $\alpha, \beta, \delta, \gamma$ refer to the interaciton coefficients. Either initial population or interaction coefficients are set to fixed in a specific mode, while another group is randomly initialized. All the values have been normalized to avoid the impact caused by magnitude.}
    \label{tab:2d_dataset}
\end{table}

\subsection{Higher-dimensional time-series}
We experiment with two higher-dimensional time-series dataset, where observations are depicted as images, including a BouncingBall dataset and a RotMNIST dataset.
For Bouncing Ball, each trajectory contains the movements of three interacting balls within a rectangular box, with the length of 20 steps, where each timestep is framed as a $32 * 32$ image.
We randomly grab 10000 abd 500 trajectories for training and testing, respectively.
For RotMNIST, each trajectory contains the rotation of a handwritten digit "3" presenting 16 angles (so as with the length of 16 steps), where each timestep is frame as a $28 * 28$ image.
Randomly drawn 400 sequences are used for experiment with a 9:1 ratio for splitting training/testing set.
During the training phase for both datasets, the context size and extra target size are randomly drawn in the range $(0, 5]$ and $(0, 5]$.
In this case, the dimension of observation space are $\mathcal{Y} \subseteq \mathbb{R}^{784}$ (RotMNIST) and $\mathcal{Y} \subseteq \mathbb{R}^{1024}$ (Bouncing Ball).

\begin{figure}[!htbp]
    \centering
    \includegraphics[width=\linewidth]{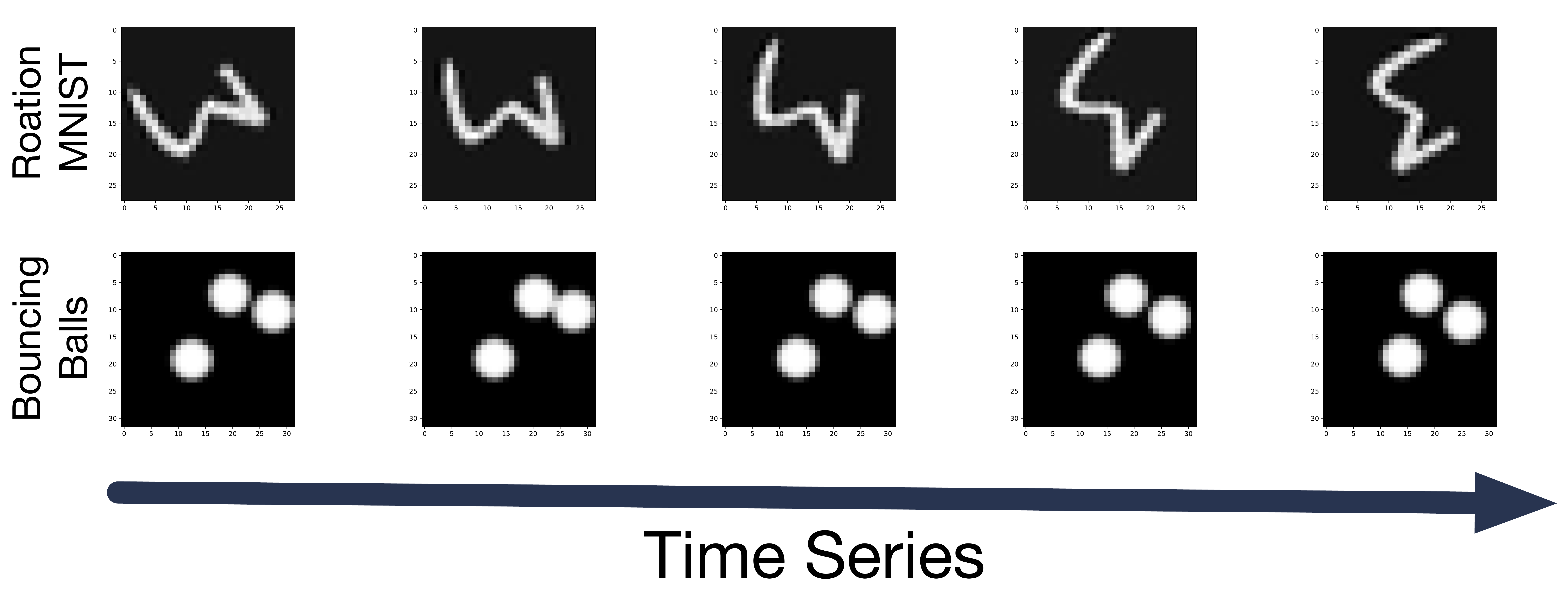}
    \caption{Examples of 5 consecutive steps of Bouncing Ball data and RotMNIST data.}
\end{figure}

\section{Additional Experimental Results}

\subsection{Few-shot Regression on GP-generated Function}
We supplement the results for 5-shot and 20-shot regression for GP-generated functions in three different kernels, with reconstruction error (MSE) used as the evaluation metrics.
It is noticeable that ConvCNP shows particular merits when predicting periodic data, while CCNP performs better with noise contained (Noisy Mat\'ern).
Also, the {\it translation-equivariance} assumption baked into ConvCNP may not hold for every case.

\begin{table}[h]
    \centering
    \resizebox{\linewidth}{!}{
        \begin{tabular}{c|c|c|c|c|c|c}
            \toprule
                    & \multicolumn{3}{c|}{{\bf 5-shot regression}} & \multicolumn{3}{c}{{\bf 20-shot regression}}                                                                                                                                 \\
            \midrule
                    & {\bf RBF}                                    & {\bf Periodic}                               & {\bf Mat\'ern}                & {\bf RBF}                     & {\bf Periodic}                & {\bf Mat\'ern}                \\
            \midrule
            CNP     & 0.723 $\pm$ 0.008                            & 0.588 $\pm$ 0.001                            & 0.972 $\pm$ 0.005             & 0.456 $\pm$ 0.019             & 0.535 $\pm$ 0.001             & 0.923 $\pm$ 0.044             \\
            AttnCNP & 0.587 $\pm$ 0.004                            & 0.552 $\pm$ 0.007                            & 0.908 $\pm$ 0.007             & 0.119 $\pm$ 0.003             & 0.497 $\pm$ 0.022             & 0.569 $\pm$ 0.283             \\
            ConvCNP & \underline{0.581 $\pm$ 0.004}                & {\bf 0.278 $\pm$ 0.007}                      & \underline{0.785 $\pm$ 0.013} & \underline{0.102 $\pm$ 0.001} & {\bf 0.065 $\pm$ 0.128}       & \underline{0.468 $\pm$ 0.151} \\
            CCNP    & {\bf 0.509 $\pm$ 0.032}                      & \underline{0.443 $\pm$ 0.138}                & {\bf 0.635 $\pm$ 0.004}       & {\bf 0.100 $\pm$ 0.015}       & \underline{0.183 $\pm$ 0.029} & {\bf 0.412 $\pm$ 0.002}       \\
            \bottomrule
        \end{tabular}
    }
    \caption{Additional 1D Few-Shot Regression with GP in three kernels, running over 3 different seeds.}
\end{table}

\subsection{Running Efficiency}
\begin{figure}
    \centering
    \includegraphics[width=0.8\linewidth]{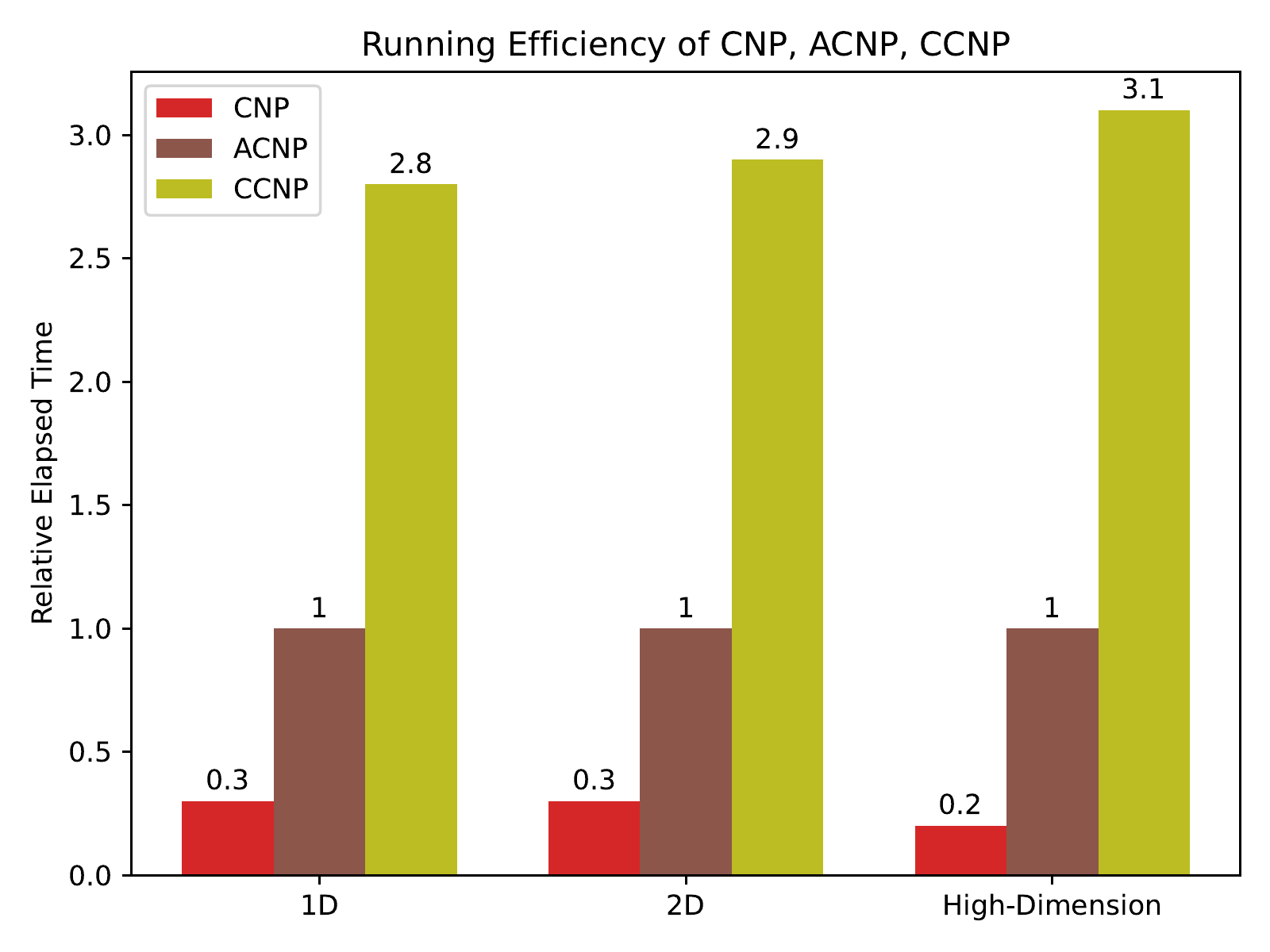}
    \caption{Comparison on Computational Efficiency}
    \label{fig: app_efficiency}
\end{figure}
We have discussed that one of the limitations of CCNP is running efficiency due to the negative sampling steps when performing contrastive learning.
We provide a comparison of initial relative elapsed time between CNP, AttnCNP, and CCNP running on the same epochs (see \cref{fig: app_efficiency}).
CCNP primarily emphasizes predictive efficacy, so its efficiency may be sacrificed somewhat.
Meanwhile, AttnCNP and CCNP are more comparable since both involve calculating attention, while CCNP costs more time to train contrastive targets.
It might be possible to optimize CCNP's efficiency by using ideas like MoCo.

\subsection{Ablation studies for Projection Heads}
As it is a commonsense that the projection head plays a significant role in determining the performance of contrastive learning, we also examine the effects of setting different sizes for projection head through studies performed within high-dimensional sequences.

\begin{table}[h]
    \centering
    \resizebox{0.7\linewidth}{!}{
        \begin{tabular}{c|c|c}
            \toprule
            {\bf Dim.} & {\bf RotMNIST}                & {\bf BouncingBall}            \\
                       & MSE ($\times 10^{-2}$)        & MSE ($\times 10^{-1}$)        \\
            \midrule
            8          & 0.751 $\pm$ 0.126             & 0.537 $\pm$ 0.002             \\
            16         & 0.687 $\pm$ 0.078             & 0.511 $\pm$ 0.001             \\
            32         & 0.654 $\pm$ 0.072             & 0.483 $\pm$ 0.008             \\
            64         & \underline{0.648 $\pm$ 0.044} & \underline{0.470 $\pm$ 0.005} \\
            128        & {\bf 0.646 $\pm$ 0.079}       & {\bf 0.458 $\pm$ 0.004}       \\
            \bottomrule
        \end{tabular}\label{tab: proj}
    }
    \caption{Ablation studies of Projection Head with pred\_size=10}
\end{table}

\section{Additional Implementation Details}
\label{sec: implementation}

We implement the model with PyTorch 1.8.0 on a Nvidia GTX Titan XP GPU.
See below for the details of CCNP's components.

\subsection{CCNP for 1D}
\noindent {\bf Input Encoder architecture for 1D}. \footnote{FCL(d) = Fully Connected Layer(output\_dimension)}
\begin{align*}
    (x, \bs{y}) \in (\mathbb{R}, \mathbb{R}) & \rightarrow \mathrm{FCL}(64) \rightarrow \mrm{ReLU}                              \\
                                             & \rightarrow \mathrm{FCL}(64) \rightarrow \mrm{ReLU}                              \\
                                             & \rightarrow \mathrm{FCL}(64) \rightarrow \mrm{ReLU}                              \\
                                             & \rightarrow \mathrm{FCL}(64) \Rightarrow \mbf{r}_{C} / \mbf{r}_{T} / \mbf{r}_{F}
\end{align*}

\noindent {\bf Position-Aware Self-Attention architecture for 1D}.
\begin{align*}
    \mrm{MultiHeadAttention} & = \{ K, Q, V, A, H \}                           \\
    \mrm{with} \; K          & = \mathrm{FCL}(64), \mrm{key\_transformation}   \\
    Q                        & = \mathrm{FCL}(64), \mrm{query\_transformation} \\
    V                        & = \mathrm{FCL}(64), \mrm{value\_transformation} \\
    A                        & = \mrm{DotProductAttention}(K, Q, V)            \\
    H                        & = \mrm{FCL}(64), \mrm{head\_fusion}
\end{align*}
where
\begin{align*}
    \mrm{DotProductAttention} & = \{ K, Q, V \}                                 \\
    \mrm{with} \; K           & = \mathrm{FCL}(64), \mrm{key\_transformation}   \\
    Q                         & = \mathrm{FCL}(64), \mrm{query\_transformation} \\
    V                         & = \mathrm{FCL}(64), \mrm{value\_transformation}
\end{align*}
\noindent {\bf Temporal Contrastive Component for 1D}.
\begin{align*}
    (x_{t}, \mbf{r}_{T}) \in (\mathbb{R}, \mathbb{R}^{64}) & \rightarrow \mrm{FCL}(64) \rightarrow \mrm{ReLU}                \\
                                                           & \rightarrow \mrm{FCL}(64) \Rightarrow \varphi(x_{t}, \mbf{r}_T) \\
    \varphi(x_{t}, \mbf{r}_T)                              & \rightarrow \mrm{FCL}(8) \Rightarrow \hat{\bs{z}}_{t}           \\
    \bs{y}_{t}                                             & \rightarrow \mrm{FCL}(8) \Rightarrow \bs{z}_{t}                 \\
    (\hat{\bs{z}}_{t}, \bs{z}_{t})                         & \rightarrow \mrm{InfoNCE}                                       \\
    \mrm{with} \; \tau                                     & = 0.5
\end{align*}

\noindent {\bf Function Contrastive Component for 1D}.
Taking 2 instantiations for illustration.
We use instantiations within a whole batch in practice.
\begin{align*}
    \mbf{x}_{C}^{f_{1}}, \mbf{x}_{C}^{f_{2}}                                                           & \rightarrow \mbf{x}_{C_{1}}^{f_{1}}, \mbf{x}_{C_{2}}^{f_{1}}, \mbf{x}_{C_{1}}^{f_{2}}, \mbf{x}_{C_{2}}^{f_{2}} \\
    \mbf{x}_{C_{1}}^{f_{1}}, \mbf{x}_{C_{2}}^{f_{1}}, \mbf{x}_{C_{1}}^{f_{2}}, \mbf{x}_{C_{2}}^{f_{2}} & \rightarrow \mbf{r}_{C_{1}}^{f_{1}}, \mbf{r}_{C_{2}}^{f_{1}}, \mbf{r}_{C_{1}}^{f_{2}}, \mbf{r}_{C_{2}}^{f_{2}} \\
    \repf{C_{1}}{f_{1}}                                                                                & \rightarrow \mrm{FCL}(8) \Rightarrow \mbf{q}_{i}^{f_{1}}                                                       \\
    \repf{C_{2}}{f_{1}}                                                                                & \rightarrow \mrm{FCL}(8) \Rightarrow \mbf{q}_{j}^{f_{1}}                                                       \\
    \repf{C_{1}}{f_{2}}                                                                                & \rightarrow \mrm{FCL}(8) \Rightarrow \mbf{q}_{i}^{f_{2}}                                                       \\
    \repf{C_{2}}{f_{2}}                                                                                & \rightarrow \mrm{FCL}(8) \Rightarrow \mbf{q}_{j}^{f_{2}}                                                       \\
    \mbf{q}_{i}^{f_{1}}, \mbf{q}_{j}^{f_{1}}, \mbf{q}_{i}^{f_{2}}, \mbf{q}_{j}^{f_{2}}                 & \rightarrow \mrm{InfoNCE}                                                                                      \\
    \mrm{with} \; \tau                                                                                 & = 0.5
\end{align*}

\noindent {\bf Output Decoder architecture for 1D}.
\begin{align*}
    (x_{t}, \mbf{r}_{C}, \mbf{r}_{T}, \mbf{r}_{F}) \in (\mathbb{R}, \mathbb{R}^{64}, \mathbb{R}^{64}, \mathbb{R}^{64}) & \rightarrow \mrm{concat}(\cdot, \cdot, \cdot, \cdot)  \\
                                                                                                                       & \rightarrow \mrm{FCL}(64) \rightarrow \mrm{ReLU}      \\
                                                                                                                       & \rightarrow \mrm{FCL}(64) \rightarrow \mrm{ReLU}      \\
                                                                                                                       & \rightarrow \mrm{FCL}(64) \rightarrow \mrm{ReLU}      \\
                                                                                                                       & \rightarrow \mrm{FCL}(64) \rightarrow \mrm{ReLU}      \\
                                                                                                                       & \Rightarrow \mbf{r}_{g}                               \\
    \mbf{r}_{g} \in \mathbb{R}^{64}                                                                                    & \rightarrow \mrm{FCL}(1) \Rightarrow \hat{\mu}_{t}    \\
    \mbf{r}_{g} \in \mathbb{R}^{64}                                                                                    & \rightarrow \mrm{FCL}(1) \Rightarrow \hat{\sigma}_{t}
\end{align*}

\subsection{CCNP for 2D}
Similar to 1D, except for $\bs{y} \in \mathbb{R}^{2}$, thus in decoder
\begin{align*}
    \mbf{r}_{g} \in \mathbb{R}^{64} & \rightarrow \mrm{FCL}(2) \Rightarrow \hat{\bs{\mu}}_{t}    \\
    \mbf{r}_{g} \in \mathbb{R}^{64} & \rightarrow \mrm{FCL}(2) \Rightarrow \hat{\bs{\sigma}}_{t}
\end{align*}

\subsection{CCNP for High-dimensional data}
We replace $\psi (\cdot)$ for encoding observations $\mbf{y}$ with Convolutional Blocks~\footnote{Conv(f, k, s, p) = Convolution2D(feat\_maps, kernel, stride, pad)}, with BN\footnote{BN(d) = BatchNormalization2D(dim)}.
For decoding we use ConvT\footnote{ConvT(f, k, s, p) = ConvTranspose2D(feat\_map, kernel, stride, pad)}

\noindent {\bf Input Encoder architecture for RotMNIST}.
\begin{align*}
    \bs{y} \in \mathbb{R}^{784}      & \rightarrow \mathrm{Conv}(16, 5, 2, 2) \rightarrow \mrm{BN}(16) \rightarrow \mrm{ReLU}   \\
                                     & \rightarrow \mathrm{Conv}(32, 5, 2, 2) \rightarrow \mrm{BN}(32) \rightarrow \mrm{ReLU}   \\
                                     & \rightarrow \mathrm{Conv}(64, 5, 2, 2) \rightarrow \mrm{BN}(64) \rightarrow \mrm{ReLU}   \\
                                     & \rightarrow \mathrm{Conv}(128, 5, 2, 2) \rightarrow \mrm{BN}(128) \rightarrow \mrm{ReLU} \\
    (x \in \mathbb{R}, \psi(\bs{y})) & \rightarrow \mathrm{FCL}(128) \Rightarrow \mbf{r}_{C} / \mbf{r}_{T} / \mbf{r}_{F}
\end{align*}

\noindent {\bf Output Decoder architecture for RotMNIST}.
\begin{align*}
    (x_{t}, \mbf{r}_{C}, \mbf{r}_{T}, \mbf{r}_{F}) \in (\mathbb{R}, \mathbb{R}^{128}, \mathbb{R}^{128}, \mathbb{R}^{128}) & \rightarrow \mrm{concat}(\cdot, \cdot, \cdot, \cdot) \\
                                                                                                                          & \rightarrow \mrm{FCL}(72) \Rightarrow \mbf{r}_{g}
\end{align*}
\begin{align*}
    \mbf{r}_{g} & \rightarrow \mrm{reshape} (\mrm{batch\_size}, 8, 28, 28)                                        \\
                & \rightarrow \mathrm{ConvT}(128, 3, 1, 0) \rightarrow \mrm{BN}(128) \rightarrow \mrm{ReLU}       \\
                & \rightarrow \mathrm{ConvT}(64, 5, 2, 0) \rightarrow \mrm{BN}(64) \rightarrow \mrm{ReLU}         \\
                & \rightarrow \mathrm{ConvT}(32, 5, 2, 1) \rightarrow \mrm{BN}(32) \rightarrow \mrm{ReLU}         \\
                & \rightarrow \mathrm{ConvT}(1, 5, 1, 2) \Rightarrow \hat{\bs{\mu}} \in \mathbb{R}^{28 \times 28} \\
\end{align*}

\noindent {\bf Input Encoder architecture for Bouncing Ball}.
\begin{align*}
    \bs{y} \in \mathbb{R}^{1024}     & \rightarrow \mathrm{Conv}(16, 5, 2, 2) \rightarrow \mrm{BN}(16) \rightarrow \mrm{ReLU}   \\
                                     & \rightarrow \mathrm{Conv}(32, 5, 2, 2) \rightarrow \mrm{BN}(32) \rightarrow \mrm{ReLU}   \\
                                     & \rightarrow \mathrm{Conv}(64, 5, 2, 2) \rightarrow \mrm{BN}(64) \rightarrow \mrm{ReLU}   \\
                                     & \rightarrow \mathrm{Conv}(128, 5, 2, 2) \rightarrow \mrm{BN}(128) \rightarrow \mrm{ReLU} \\
    (x \in \mathbb{R}, \psi(\bs{y})) & \rightarrow \mathrm{FCL}(128) \Rightarrow \mbf{r}_{C} / \mbf{r}_{T} / \mbf{r}_{F}
\end{align*}

\noindent {\bf Output Decoder architecture for Bouncing Ball}.
\begin{align*}
    (x_{t}, \mbf{r}_{C}, \mbf{r}_{T}, \mbf{r}_{F}) \in (\mathbb{R}, \mathbb{R}^{128}, \mathbb{R}^{128}, \mathbb{R}^{128}) & \rightarrow \mrm{concat}(\cdot, \cdot, \cdot, \cdot) \\
                                                                                                                          & \rightarrow \mrm{FCL}(72) \Rightarrow \mbf{r}_{g}
\end{align*}
\begin{align*}
    \mbf{r}_{g} & \rightarrow \mrm{reshape} (\mrm{batch\_size}, 8, 32, 32)                                        \\
                & \rightarrow \mathrm{ConvT}(128, 3, 2, 1) \rightarrow \mrm{BN}(128) \rightarrow \mrm{ReLU}       \\
                & \rightarrow \mathrm{ConvT}(64, 5, 2, 1) \rightarrow \mrm{BN}(64) \rightarrow \mrm{ReLU}         \\
                & \rightarrow \mathrm{ConvT}(32, 5, 2, 1) \rightarrow \mrm{BN}(32) \rightarrow \mrm{ReLU}         \\
                & \rightarrow \mathrm{ConvT}(1, 5, 1, 2) \Rightarrow \hat{\bs{\mu}} \in \mathbb{R}^{32 \times 32} \\
\end{align*}

{\small
\bibliographystyle{ieee_fullname}
\bibliography{egbib}
}